%% file: 0_main.tex
\documentclass[letterpaper]{article} 
\usepackage{aaai2026}  
\usepackage{times}  
\usepackage{helvet}  
\usepackage{courier}  
\usepackage[hyphens]{url}  
\usepackage{graphicx} 
\usepackage{natbib}  
\usepackage{caption} 
\usepackage{algorithm}
\usepackage{algorithmic}

\usepackage{newfloat}
\usepackage{listings}
\DeclareCaptionStyle{ruled}{labelfont=normalfont,labelsep=colon,strut=off} 
\floatstyle{ruled}
\newfloat{listing}{tb}{lst}{}
\floatname{listing}{Listing}
\usepackage{bibentry}

\usepackage{algorithm}
\usepackage{algorithmic}
\usepackage{amsmath}
\usepackage{amsfonts}
\usepackage{booktabs}
\usepackage{makecell}
\usepackage{array} 
\usepackage{multirow}
\usepackage{pifont}
\usepackage{xcolor}
\usepackage{subcaption}
\usepackage{appendix}
\newcounter{challnum}
\newcounter{insightnum}
\usepackage{xspace}
\usepackage{longtable}
\usepackage{tabularx}
\usepackage{rotating}
\usepackage{pdflscape}
\usepackage{subcaption}
\usepackage{fancyhdr}
\usepackage{lastpage}
\newcommand{\mychall}{%
  \noindent\refstepcounter{challnum}\thechallnum%
}
\newcommand{\myinsight}{%
  \noindent\refstepcounter{insightnum}\theinsightnum%
}

\usepackage{newfloat}
\usepackage{listings}
\DeclareCaptionStyle{ruled}{labelfont=normalfont,labelsep=colon,strut=off} 
\floatstyle{ruled}
\newfloat{listing}{tb}{lst}{}
\floatname{listing}{Listing}
\usepackage[skins]{tcolorbox}
\tcbuselibrary{listingsutf8}
\usepackage[labelformat=simple]{subcaption}
\global

\tcbset{
  aibox/.style={
    width=\columnwidth,              
    enhanced,
    colback=white,
    colframe=black!70,
    colbacktitle=black!5,
    coltitle=black,
    fonttitle=\bfseries,
    boxsep=3pt,
    left=6pt,right=6pt,top=6pt,bottom=6pt,
    fontupper=\small,
    attach boxed title to top left={xshift=0.6ex,yshift=-0.6ex},
    boxed title style={boxrule=0pt,colframe=white},
    halign=flush left,
    halign title=flush left,
    before=\par\noindent,           
    before upper=\raggedright       
  }
}
\newtcolorbox{AIbox}[2][]{aibox,title=#2,#1}

\title{
iMAD: Intelligent Multi-Agent Debate for Efficient and Accurate LLM Inference
}
\author{
    Wei Fan,
    JinYi Yoon,
    Bo Ji
}
\affiliations{
    Department of Computer Science, Virginia Tech, Blacksburg, VA, USA\\
    \{fanwei, jinyiyoon, boji\}@vt.edu}

\usepackage{times}
\usepackage{helvet}
\usepackage{courier}
\usepackage{xcolor}
\newcounter{checksubsection}
\newcounter{checkitem}[checksubsection]

\newcommand{\name}{\text{iMAD}\xspace}
\begin{document}

\maketitle

\input{1_abstract}
\input{2_Introduction}

\input{3_Related_Work}
\input{4_Key_Insights}
\input{5_Our_Design}
\input{6_Evaluation}
\input{8_Conclusion}
\input{9_Acknowledgement}

\bibliography{aaai2026}
\input{10_Appendix}

\end{document}

%% file: 1_abstract.tex
\begin{abstract}
Large Language Model (LLM) agent systems have advanced rapidly, driven by their strong generalization in zero-shot settings. To further enhance reasoning and accuracy on complex tasks, Multi-Agent Debate (MAD) has emerged as a promising framework that engages multiple LLM agents in structured debates to encourage diverse reasoning. However, triggering MAD for every query is inefficient, as it incurs substantial computational (token) cost and may even degrade accuracy by overturning correct answers from single-agent. To address these limitations, we propose intelligent Multi-Agent Debate (\name), a token-efficient framework that selectively triggers MAD only when it is likely to be beneficial (i.e., correcting an initially wrong answer). To achieve this goal, \name learns generalizable model behaviors to make accurate debate decisions. Specifically, \name first prompts a single agent to produce a structured self-critique response, from which we extract 41 interpretable linguistic and semantic features capturing hesitation cues. Then, \name uses a lightweight debate decision classifier, trained using our proposed FocusCal loss without test-dataset-specific tuning, to make robust zero-shot debate decisions. Through extensive experiments using six (visual) question answering datasets against five competitive baselines, we show that \name significantly reduces token usage (by up to 92\%) while also improving final answer accuracy (by up to 13.5\%).
\end{abstract}

\begin{links}
    \link{Code}{https://github.com/Fanwei100/iMAD}
\end{links}

%% file: 2_Introduction.tex
    \begin{figure}[t]
    \centering
    \includegraphics[width=0.92\linewidth]{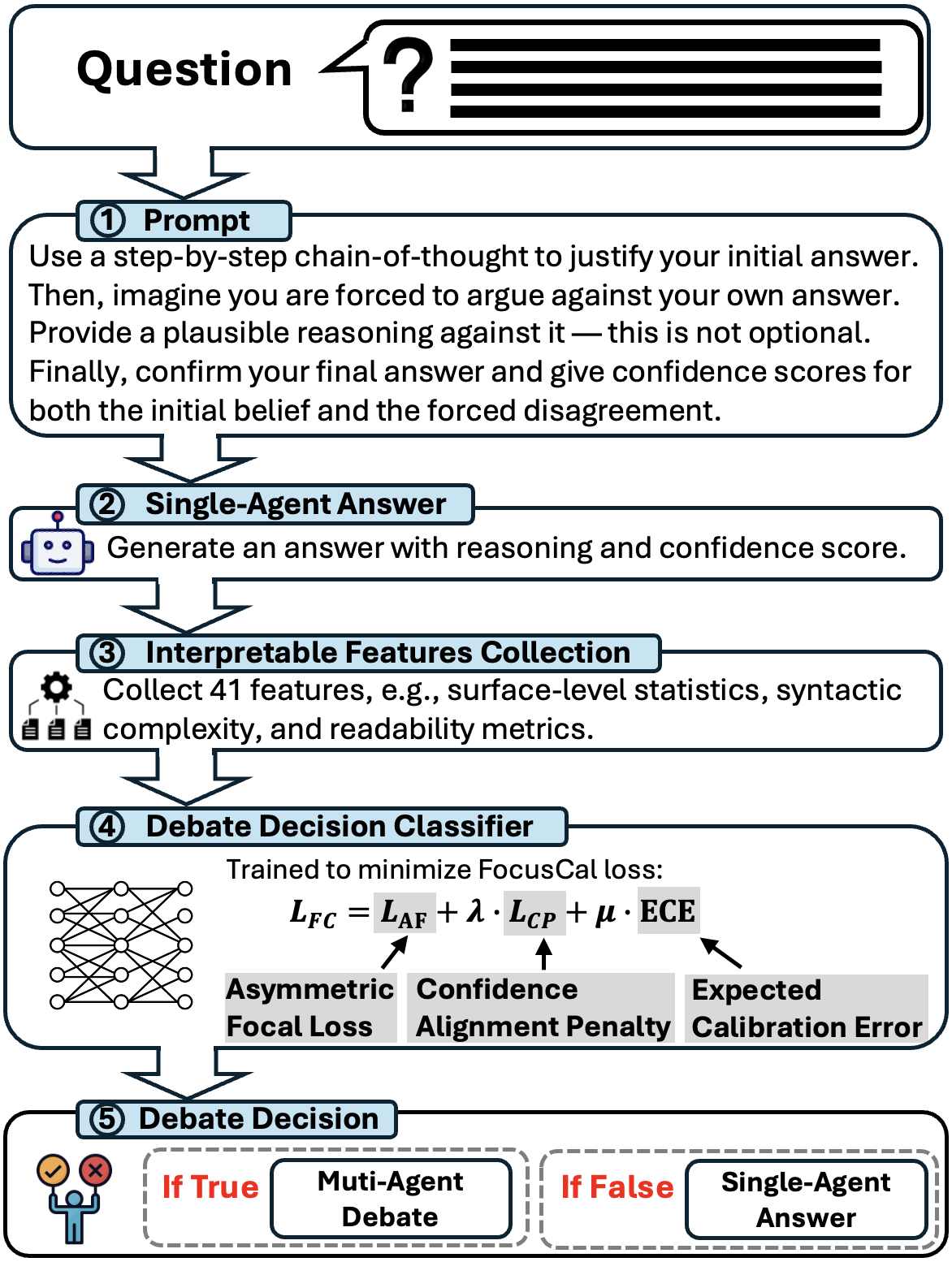}
    \caption{Overall workflow of \name.} 
    \label{fig:fmad_workflow}
    \end{figure}    
\section{Introduction}
With the rapid progress in Large Language Models (LLMs), agent systems have shown impressive zero-shot reasoning capabilities across tasks such as (visual) question answering, problem solving, or code generation. This ability in zero-shot settings, without access to evaluation data, makes LLM agents appealing for real-world applications by enabling fast and scalable deployment across diverse domains~\cite{wan-etal-2023-better, verbalizedconfidencescoresllms}. These agent systems typically rely on a single LLM agent to generate step-by-step reasoning, using methods like Chain-of-Thought (CoT)~\cite{wei2022chain} or Self-Consistency~\cite{wang2022self}. However, these approaches often suffer from limited diversity and may overlook alternative reasoning paths as they rely on a single agent's perspective. 

To address this limitation, recent studies have explored multi-agent systems that enable collaboration or debate among multiple agents to enhance reasoning and decision-making~\cite{liang-etal-2024-encouraging, chen2023reconcile}. Among these approaches, \emph{Multi-Agent Debate (MAD)}, inspired by the role of structured discourse in human cognition, has emerged as a particularly promising framework~\cite{wu2023autogenenablingnextgenllm}. In MAD, multiple agents independently reason over a query and critique each other’s output through structured interactions, stimulating adversarial dialogue and iterative refinement~\cite{liang-etal-2024-encouraging,tillmann2025literaturereviewmultiagentdebate}. Such interactions encourage diverse reasoning paths and perspective shifts, enabling agents to recover from faulty initial answers and thus often outperform single-agent systems~\cite{liang-etal-2024-encouraging, hsu2025groupthinkmultipleconcurrent}. 

Despite this benefit, MAD frameworks face two critical limitations that hinder their practical deployment. First, \emph{due to iterative LLM queries, MAD incurs significantly higher computational costs}--measured in total token usage, which includes both input tokens (prompts to each agent) and output tokens (generated responses). Most MAD systems consume three to five times more tokens than single-agent baselines~\cite{liu2024groupdebateenhancingefficiencymultiagent}, making them costly to scale (see Insight~\ref{insight:first}). 
Second, perhaps more counterintuitively, \emph{MAD does not consistently improve response quality}. 
Both prior work~\cite{zhang2025stopovervaluingmultiagentdebate} and our own empirical analysis show that in many cases, the single-agent output is already correct, making it redundant to trigger MAD. In other cases, the error in the single-agent response cannot be corrected by MAD, or even worse, triggering MAD may override a correct answer with an incorrect one, resulting in degraded accuracy (see Insight~\ref{insight:second}). These observations suggest that while MAD can be beneficial, applying it to every query may be inefficient. Not only does it incur substantial computational costs, but it can also degrade accuracy. Therefore, we need an intelligent and principled mechanism that can selectively trigger MAD only when it is likely to be beneficial. This raises a key question: \emph{When should a debate be performed to preserve the benefits of MAD while avoiding unnecessary token costs and potential accuracy degradation?}

A straightforward approach is to use the confidence score (typically computed as the average log-probability of the output tokens) to estimate single-agent answer correctness. Leveraging this idea, a concurrent work recently proposed a selective MAD framework, called DOWN, which attempts to trigger MAD when the confidence score falls below a threshold~\cite{eo2025debatenecessaryadaptivemultiagent}. However, it requires a threshold tuned on a subset of the evaluation data, which violates the core zero-shot setting assumption. Moreover, even with a fixed threshold, our empirical analysis reveals that confidence scores alone are not reliable indicators of whether MAD is necessary (see Insight~\ref{insight:third}). We observe that confidence scores could be high even for incorrect answers, revealing the model’s overconfidence. Also, confidence scores are often misaligned with reasoning uncertainty: responses that contain hesitation cues (such as hedging, contradictions, or shallow reasoning) may still receive a high score. This misalignment causes two undesirable situations: skipping the necessary MAD or triggering it unnecessarily.

While existing methods like repeated sampling (i.e., querying the same agent multiple times) or invoking more LLM agents can estimate whether MAD is necessary, they incur high token costs and thus are not scalable~\cite{wang2022self}. 
This highlights the need for token-efficient mechanisms that can make informed decisions about when to trigger MAD based on the initial single-agent output. There are two key challenges to achieving this goal:

(C\mychall\label{chall:first}) \emph{How to design an effective prompt that guides the single-agent response to expose richer features for making a debate decision?} This decision must rely on features embedded in the initial single-agent output. These features need to be informative and can be efficiently extracted (i.e., without requiring repeated sampling or additional LLM queries).

(C\mychall\label{chall:second}) \emph{How to design a mechanism that intelligently decides when to trigger MAD for efficient and accurate inference in the zero-shot setting?}
One needs to identify when MAD is more likely to recover incorrect answers without relying on the evaluation dataset. This requires learning generalizable model behaviors, which involves addressing the aforementioned issues of LLM agents: overconfidence in incorrect answers and misalignment between confidence scores and semantic cues of hesitation in the response.

To address these challenges, we propose \underline{\textbf{i}}ntelligent \underline{\textbf{M}}ulti-\underline{\textbf{A}}gent \underline{\textbf{D}}ebate (\name), a lightweight debate-triggering framework for token-efficient MAD in the zero-shot setting. \name selectively triggers MAD only when it is more likely to improve the final answer. The overall workflow of \name is illustrated in Fig.~\ref{fig:fmad_workflow}. We summarize our main contributions along with the key components of \name as follows:
\begin{itemize}
    \item To address Challenge (C\ref{chall:first}), we propose the \name framework with a structured self-critique prompt (Steps~{\large \textcircled{\small 1}}-{\large \textcircled{\small 2}} in Fig.~\ref{fig:fmad_workflow}). This prompt directs a single agent to produce (\romannumeral1) an initial CoT justification, (\romannumeral2) a required self-critique that argues for a plausible alternative, and (\romannumeral3) confidence scores for both perspectives. This prompt stimulates a lightweight internal mini-debate without adding input tokens and incurs only minimal additional output tokens. This design offers rich semantic and uncertainty cues, enabling accurate and token-efficient debate decisions.
    
    \item To address Challenge (C\ref{chall:second}), we formulate MAD triggering as a classification problem. From each structured single-agent response, we extract 41 interpretable linguistic and semantic features along with the confidence score (Step~{\large \textcircled{\small 3}}), which will be fed into a lightweight debate decision classifier of a multi-layer perceptron (MLP) (Step~{\large \textcircled{\small 4}}). To enable accurate decisions in zero-shot settings, the classifier learns generalizable model behaviors using a 
    proposed Confidence-Calibrated \emph{FocusCal} loss that integrates: (\romannumeral1) \emph{Asymmetric Focal loss} ($L_{\text{AF}}$) to penalize overconfident errors and emphasize incorrect cases; (\romannumeral2) \emph{Confidence Penalty} ($L_{\text{CP}}$) to penalize misalignment between confidence scores and semantic uncertainty in the response; and (\romannumeral3) \emph{Expected Calibration Error} (ECE) to encourage the predicted debate-triggering score to align with empirical correctness. This enables the classifier to prioritize debatable cases (i.e., recoverable errors) while reducing unnecessary MAD (Step~{\large \textcircled{\small 5}}).
    
    \item We evaluate \name on three question answering (QA) and three visual question answering (VQA) datasets against five competitive baselines (including two single-agent and three multi-agent frameworks). \name reduces token usage by up to 92\% while improving accuracy by up to 13.5\% through selectively skipping debates that are unnecessary or detrimental. Notably, we train the classifier solely on two representative datasets selected to capture diverse model reasoning behaviors, allowing the classifier to learn generalizable model behaviors and perform effectively across six held-out datasets.
    \end{itemize}

%% file: 3_Related_Work.tex
\section{Related Work}

We categorize highly related works into three groups.

\paragraph{Single-Agent and Multi-Agent LLMs.}
LLMs have demonstrated strong reasoning capabilities in single-agent settings, where a single LLM agent performs all reasoning steps independently. Foundational methods like CoT prompt models to generate intermediate reasoning steps, enabling better handling of complex tasks~\cite{wei2022chain}. Recently, Self-Consistency further improves accuracy over CoT by sampling multiple outputs and selecting the most frequent answer~\cite{wang2022self, wang2024soft, li2024escape}. However, single-agent reasoning relies solely on internal sampling, lacking perspective diversity and explicit self-correction. To address this, multi-agent LLM frameworks leverage multiple agents to reason independently or coordinate through structured interaction, thus improving accuracy over single-agent approaches. Some methods generate multiple outputs for joint evaluation (e.g., CoMM~\cite{Chen2024CoMMCM}), while others assign hierarchical agent roles (e.g., Mixture-of-Agents (MoA) to progressively refine reasoning~\cite{wang2024moa, chen2024moa, li2025rethinking}). While these methods stimulate collaborative reasoning among agents, they often incur high computational costs and deliver inconsistent improvements over single-agent baselines~\cite{zhang2025stopovervaluingmultiagentdebate, cemri2025multi}.

\paragraph{MAD Frameworks.}
MAD frameworks represent a structured subclass of multi-agent LLM where agents engage in explicit argumentative exchanges (e.g., critiques, rebuttals, or deliberation) to refine initial outputs~\cite{tillmann2025literaturereviewmultiagentdebate}. Existing MAD methods include role-based debates with assigned affirmative, negative, and moderator roles~\cite{liang-etal-2024-encouraging, wang2024unleashingemergentcognitivesynergy}, implicit debate via input perturbation and aggregation (e.g., Reconcile~\cite{chen2023reconcile}), and intra-agent self-refinement~\cite{srivastava2025debate, zhang2024can}. Building on these MAD designs, GroupDebate extends MAD by coordinating subgroup debate~\cite{liu2024groupdebateenhancingefficiencymultiagent}. 
While MAD enhances interpretability and error correction, recent studies show that it can also introduce noise or overturn correct single-agent answers, thus degrading accuracy~\cite{zhang2025stopovervaluingmultiagentdebate}.

\paragraph{Confidence-based Selective Debate.} 
The concurrent work \textsc{DOWN} aims to reduce MAD token costs by using LLM-generated confidence scores to decide when to trigger debate~\cite{eo2025debatenecessaryadaptivemultiagent}. However, selecting an appropriate confidence threshold requires labeled evaluation data, which violates the zero-shot setting assumption normally upheld by current single-agent and MAD baselines and limits real-world applicability. Moreover, confidence scores alone are unreliable in determining whether triggering a debate will improve the answer (see Insight~\ref{insight:third}). Although DOWN saves tokens by skipping debate on high-confidence responses, it often fails to identify cases where debate is beneficial and misses opportunities to correct recoverable errors. 

%% file: 4_Key_Insights.tex
\section{Key Insights}
\label{sec:insight}
\begin{table}[t]
\centering 
\setlength{\tabcolsep}{1.8pt}
\begin{tabular}{lcccc}
\toprule
\multirow{2}{*}{\centering\textbf{Dataset}} 
& \multicolumn{2}{c}{\makecell{\textbf{Single-Agent} \\ \textbf{(CoT)}}} 
& \multicolumn{2}{c}{\makecell{\textbf{Multi-Agent} \\ \textbf{Debate} \textbf{(MAD)}}} \\
\cmidrule(lr){2-3} \cmidrule(lr){4-5}
& \textbf{Acc (\%)} & \textbf{\# Token}
& \textbf{Acc (\%)} & \textbf{\# Token} \\
\midrule
MEDQA     & 76.6 & 653  & 81.9 & 4,034 \\
MMLU      & 86.8 & 764  & 89.5 & 3,348 \\
GSM8K     & 71.3 & 618  & 76.4 & 3,446 \\
\midrule
OKVQA     & 88.3 & 1,945 & 89.8 & 7,803 \\
VQA-v2       & 77.5 & 2,245 & 81.0 & 8,796 \\
ScienceQA & 86.0 & 1,720 & 89.4 & 6,777 \\
\bottomrule
\end{tabular}
\caption{Comparison of accuracy (Acc) and average token costs (\# Token) per question between single-agent CoT and MAD frameworks across QA and VQA datasets.}
\label{tab:single_vs_multi_agent}
\end{table}
While MAD shows promise in enhancing LLM reasoning, its deployment remains limited, as modest accuracy gains often come with substantial computational cost. In this section, we first quantify the token overhead of MAD compared to single-agent CoT. Then, we analyze when MAD is beneficial and reveal that the positive impact exists only for a subset of instances. Finally, we examine whether standard uncertainty heuristics (e.g., based on confidence score) can effectively guide debate triggering.

\paragraph{(Insight \myinsight\label{insight:first}) MAD achieves a higher accuracy at the cost of a substantial token overhead.} 
We quantify the trade-off between accuracy and token costs by comparing CoT~\cite{wei2022chain} and MAD~\cite{liang-etal-2024-encouraging} across six QA and VQA datasets in Table~\ref{tab:single_vs_multi_agent}. Consistent with prior findings~\cite{liu2024groupdebateenhancingefficiencymultiagent}, we observe that MAD achieves a higher accuracy than CoT, with gains ranging from 1.5\% (on OKVQA) to 5.3\% (on MEDQA). However, MAD consumes 3--5 times more tokens than CoT, mainly due to routing the same query to multiple agents, each requiring separate input prompts and generating individual responses~\cite{chen2024optima,eo2025debatenecessaryadaptivemultiagent}. This cost is more pronounced in VQA tasks, where visual inputs further increase token usage. These observations indicate that the accuracy gains come at a high token cost, making MAD impractical to deploy at scale.

\paragraph{(Insight \myinsight\label{insight:second}) The accuracy gains in MAD are primarily driven by a subset of cases.} While a concurrent work has noted a similar observation~\cite{zhang2025stopovervaluingmultiagentdebate}, we provide a systematic breakdown to quantify the specific sources of MAD’s accuracy gains. Specifically, we categorize each input instance into four cases: (\romannumeral1) incorrect in single-agent but correct in MAD ({\color{black} \ding{55}}$\rightarrow${\color{black} \ding{51}}); (\romannumeral2) correct in single-agent but incorrect in MAD ({\color{black} \ding{51}}$\rightarrow${\color{black} \ding{55}}); (\romannumeral3) correct in both ({\color{black} \ding{51}}$\rightarrow${\color{black} \ding{51}}); and (\romannumeral4) incorrect in both ({\color{black} \ding{55}}$\rightarrow${\color{black} \ding{55}}). As shown in Table~\ref{tab:mad_flipping_stats}, the ideal scenario where MAD makes corrections ({\color{black} \ding{55}}$\rightarrow${\color{black} \ding{51}}) accounts for a small portion (e.g., 4.9\% in OKVQA to 19.1\% in GSM8K). In contrast, many debates are either redundant (i.e., single-agent answers are already correct: {\color{black} \ding{51}}$\rightarrow${\color{black} \ding{51}}), ineffective (i.e., unresolved single-agent errors: {\color{black} \ding{55}}$\rightarrow${\color{black} \ding{55}}), or harmful (i.e., flipping correct single-agent answers to incorrect: {\color{black} \ding{51}}$\rightarrow${\color{black} \ding{55}}). This shows that while MAD improves the overall accuracy, the benefit is limited to a small portion of cases. Thus, indiscriminately applying MAD to all cases could waste computational resources and even degrade accuracy.

\begin{table}[t]
\centering 
\setlength{\tabcolsep}{2.6pt}
\begin{tabular}{lcccc}
\toprule
\textbf{Dataset} 
& {\color{black} \ding{55}}$\rightarrow${\color{black} \ding{51}} \textbf{(\%)} & {\color{black} \ding{51}}$\rightarrow${\color{black} \ding{55}} \textbf{(\%)} & {\color{black} \ding{51}}$\rightarrow${\color{black} \ding{51}} \textbf{(\%)} & {\color{black} \ding{55}}$\rightarrow${\color{black} \ding{55}} \textbf{(\%)} \\ 
\midrule

MEDQA     & 11.9 & 6.6 & 70.0 & 11.5 \\
MMLU      & 6.9  & 4.2 & 82.6 & 6.3  \\
GSM8K     & 19.1 & 14.0 & 57.3 & 9.6 \\
\midrule
OKVQA     & 4.9  & 3.4 & 84.9 & 6.8  \\
VQA-v2       & 9.2  & 5.7 & 71.8 & 13.3 \\
ScienceQA & 7.9  & 4.5 & 81.5 & 6.1  \\
\bottomrule
\end{tabular}
\caption{Breakdown of MAD outcomes across datasets: percentage of cases where the MAD flipped the answer correctly ({\color{black} \ding{55}}$\rightarrow${\color{black} \ding{51}}) or wrong ({\color{black} \ding{51}}$\rightarrow${\color{black} \ding{55}}) from single-agent CoT answer, and percentage of cases where both MAD and single-agent CoT are correct ({\color{black} \ding{51}}$\rightarrow${\color{black} \ding{51}}) or wrong ({\color{black} \ding{55}}$\rightarrow${\color{black} \ding{55}}).}
\label{tab:mad_flipping_stats}
\end{table}

\begin{figure}[t]
    \centering
    \includegraphics[width=0.6\linewidth]{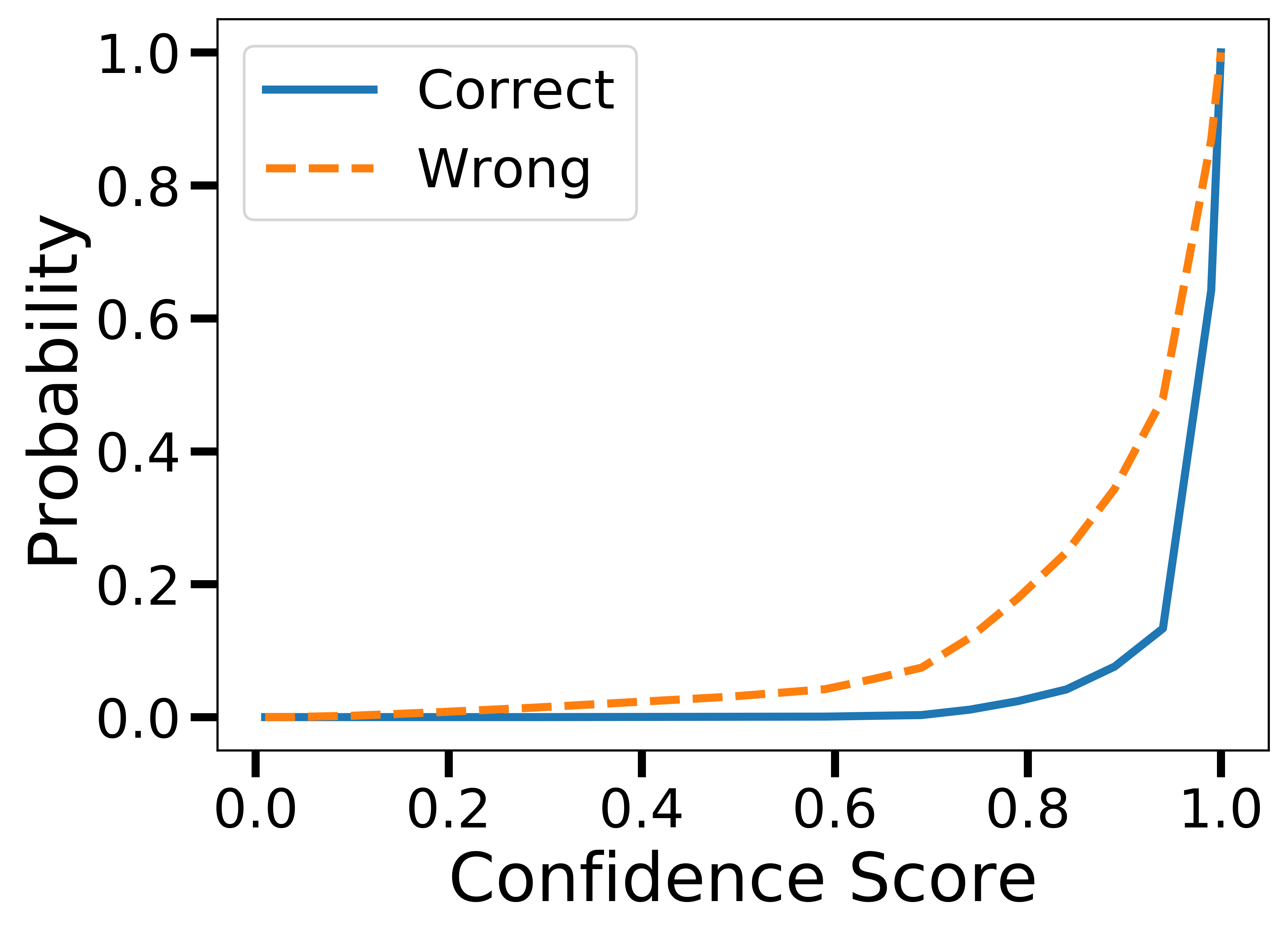}
    \caption{Cumulative Density Function (CDF) of confidence scores for correct and incorrect single-agent answers.} 
    \label{fig:PDF_Confidence}
    \end{figure}

\paragraph{(Insight \myinsight\label{insight:third}) Confidence scores are unreliable indicators of whether debate is beneficial or not.} We investigate whether LLM-generated confidence scores can effectively indicate when MAD is likely to improve answers~\cite{confidence, eo2025debatenecessaryadaptivemultiagent}. A natural hypothesis is that answers with a high confidence score may not need MAD; only answers with a low confidence score could benefit from MAD.
To test this, we consider confidence scores from various prompting strategies~\cite{confidence, eo2025debatenecessaryadaptivemultiagent} and analyze their alignment with cases where MAD corrects initially incorrect single-agent answers. In Fig.~\ref{fig:PDF_Confidence}, we observe that the Cumulative Density Function (CDF) of confidence scores is highly right-skewed and poorly aligned with answer correctness or debate effectiveness. Notably, incorrect answers often receive high confidence scores, sometimes even exceeding correct answers, indicating a strong bias of overconfidence. Moreover, even many hesitant or shallow responses still receive inflated confidence scores. This misalignment between the confidence score and the uncertainty in reasoning undermines the effectiveness of heuristics that make debate decisions based on confidence scores only. 

%% file: 5_Our_Design.tex
\section{Our Design: \name}
In this section, we present \name, a token-efficient framework that selectively triggers MAD only when it is likely to correct an initially incorrect single-agent answer. Leveraging the aforementioned insights, \name aims to substantially reduce the token overhead of MAD while retaining or even improving the accuracy (Insight~\ref{insight:first}). We begin with an overview of the \name framework (Section~\ref{sec:overview}), which integrates structured self-critique prompting with a lightweight classifier to decide whether debate should be triggered (Insight~\ref{insight:second}). The classifier leverages interpretable features extracted from the single-agent response to assess the need for debate,  enabling robust debate decisions in zero-shot settings without dataset-specific tuning. To train this classifier effectively, we propose FocusCal loss (\( L_{\text{FC}} \)) (Section~\ref{sec:FocusCal}) to address the aforementioned issues of LLM: overconfidence and misalignment between confidence scores and semantic uncertainty (Insight~\ref{insight:third}). This design enables \name to make token-efficient and accurate debate decisions based on single-agent responses in zero-shot settings.

\subsection{Framework Overview}\label{sec:overview}
As shown in Fig.~\ref{fig:fmad_workflow}, \name comprises three core stages:
(\romannumeral1) generating a structured response from a single-agent LLM using a self-critique prompt (Steps~{\large \textcircled{\small 1}}-{\large \textcircled{\small 2}});
(\romannumeral2) extracting interpretable features from the generated output (Step~{\large \textcircled{\small 3}}), and
(\romannumeral3) applying a well-trained debate decision classifier to determine whether MAD should be triggered (Steps~{\large \textcircled{\small 4}}-{\large \textcircled{\small 5}}).

\paragraph{(i) Generating Structured Self-Critique Single-Agent Response (Steps~{\large \textcircled{\small 1}}-{\large \textcircled{\small 2}}).} 
Given an input question, the system first prompts the LLM to generate a structured response with three key components:
(\romannumeral1) an initial CoT justification supporting the original answer;
(\romannumeral2) a required self-critique presenting a counterargument; and
(\romannumeral3) a final reflection including the chosen answer and explicit confidence scores for both initial reasoning and the self-critique perspectives. This structure stimulates a mini-debate: if both perspectives provide similarly strong or weak reasoning with comparable confidence scores, the model is likely to show internal hesitation, suggesting that MAD could be beneficial. Conversely, if one side presents a clear and well-supported argument while the other is weak, the answer is likely already determined, either confidently correct or confidently incorrect. In the latter case, the internally coherent but flawed reasoning makes it difficult to correct through debate.

\paragraph{(ii) Extracting Interpretable Features (Step~{\large \textcircled{\small 3}}).} We extract 41 interpretable linguistic and semantic features from the structured single-agent output, drawing from the question, initial reasoning, and self-critique. These interpretable features capture human-understandable cues to reasoning quality and internal hesitation, including surface-level statistics, readability scores, part-of-speech counts (e.g., nouns, verbs, and adjectives), question-type indicators, and lexical cues of uncertainty, such as hedging and contrast. The feature set enables fine-grained detection of subtle uncertainty cues that are often not reflected in the model’s raw confidence scores. These interpretable features help the classifier identify when MAD is likely to be beneficial by capturing uncertainty cues that are not well aligned with confidence scores. Rather than relying on a subset of features, we combine complementary semantic, syntactic, and pragmatic signals to form a holistic view of model behavior. This is crucial for generalization in zero-shot settings without access to an evaluation dataset. A detailed list of all 41 features is provided in Appendix~\ref{sec:features}.

\paragraph{(iii) Making Debate Decisions via the Classifier (Steps~{\large \textcircled{\small 4}}-{\large \textcircled{\small 5}}).}
We develop a lightweight MLP-based classifier to decide whether to trigger MAD based on features extracted from a structured single-agent response. The input feature vector \( \mathbf{z} \in \mathbb{R}^d \), a \( d \)-dimensional real-valued vector space, where
\( d \) is the total number of extracted features. The vector \( \mathbf{z}\) includes LLM-generated confidence score \( p_{\text{LLM}} \), along with semantic and linguistic features from the question, answer, and self-critique. The classifier outputs a scalar \( p \in (0,1) \), indicating the likelihood that the single-agent answer is either correct or unrecoverably incorrect (where a debate is considered unnecessary). 

During training, we run the single-agent pipeline on each instance to generate an answer and self-critique, and then assign a binary label $y$ where $y=1$ if the answer matches the ground truth and $y=0$ otherwise. The classifier is trained with these binary correctness labels, but its goal is not to replicate these labels. Instead, it aims to identify debatable cases: answers likely to be wrong but potentially correctable through MAD. This distinction is crucial because confidently incorrect answers, often resulting from coherent but flawed reasoning, are unlikely recoverable via MAD. 

During inference, we apply a decision threshold \( \tau \in (0,1) \), tuned on a validation set, to the predicted score \( p \): if \( p < \tau \), MAD is triggered; otherwise, the original single-agent answer is retained. A high \(p\) indicates triggering MAD is unnecessary, while a low \(p\) flags uncertain and potentially recoverable errors that warrant MAD. This selective mechanism enables \name to devote computational resources to the most uncertain and error-prone cases, reducing token costs while improving accuracy. The classifier is trained using our proposed FocusCal loss, detailed in Section~\ref{sec:FocusCal}.

\subsection{Debate Decision Classifier with FocusCal Loss}
\label{sec:FocusCal}
To train the classifier, we propose FocusCal loss (\( L_{\text{FC}} \)), a composite objective addressing overconfidence and uncertainty in single-agent confidence scores (Insight~\ref{insight:third}). It combines three components: (\romannumeral1) Asymmetric Focal Loss (\(L_{\text{AF}}\)) that targets the overconfidence issue by penalizing confidently incorrect predictions more than correct ones, encouraging the model to remain cautious on borderline cases; (\romannumeral2) Confidence Penalty (\( L_{\text{CP}} \)) that aligns the predicted score \(p\) with an auxiliary uncertainty score \(u \in (0,1)\), derived from semantic hesitation features via an MLP, penalizing overconfident predictions with uncertain reasoning; (\romannumeral3) Expected Calibration Error (ECE) that regularizes predicted scores for empirical calibration~\cite{Nixon_2019_CVPR_Workshops}. Together, these components help the classifier detect recoverable errors that merit MAD and avoid unnecessary debate.

To realize this design, the classifier passes the input feature vector \( \mathbf{z}\) through a shared MLP feature encoder \( f_{\text{e}}(\cdot) \) to produce a high-level representation, which is then fed into two separate output heads: a correctness head \(f_p(\cdot)\) and a hesitation head \(f_u(\cdot)\), producing two scalar logits $\ell_{\text{p}}$ and $\ell_{\text{u}}$:
\begin{equation}
    \ell_{\text{p}} := f_p(f_{\text{e}}(\mathbf{z}))~\text{and}~\ell_{\text{u}} := f_u(f_{\text{e}}(\mathbf{z})).
\end{equation}
To integrate the scalar LLM confidence score \( p_{\text{LLM}} \) with the MLP-produced logit $\ell_{\text{p}}$ and keep them mathematically consistent, we convert \( p_{\text{LLM}} \) into $\ell_{\text{LLM}}$ in the logit space:
\begin{equation}
    \ell_{\text{LLM}} := \log\left( \frac{p_{\text{LLM}}}{1 - p_{\text{LLM}}} \right).
\end{equation}
The predicted score \(p\) and the uncertainty score \(u\) are then calculated as:
\begin{align}
    p &:= \sigma(w_1 \cdot \ell_{\text{LLM}} + w_2 \cdot \ell_{\text{p}} + \epsilon), \\
    u &:= \sigma(\ell_{\text{u}}),
\end{align}
where \( w_1, w_2, \epsilon \in \mathbb{R} \) are learnable parameters and \( \sigma(\cdot) \) is the sigmoid function. This two-headed MLP separates the predicted score \(p\), used for debate decisions, from the auxiliary uncertainty score \(u\), which captures hesitation cues in the reasoning path and is supervised via \( L_{\text{CP}} \). This design enables the classifier to better identify hesitant and potentially recoverable errors that warrant debate, while skipping MAD for confidently correct answers or unrecoverable errors. 

To optimize the classifier for accurate and calibrated debate triggering, we train it using the FocusCal loss:
\begin{equation}
    L_{\text{FC}} (y, p, u) := L_{\text{AF}}(y, p) + \lambda \cdot L_{\text{CP}}(y, p, u) + \mu \cdot \text{ECE}(y, p).
\end{equation}
The weights \( \lambda\) and \(\mu \) are non-negative coefficients tuned via grid search on a held-out validation set to balance the contributions of uncertainty alignment and calibration. We discuss the details of each term in \(L_{\text{FC}}\) below.

\paragraph{Asymmetric Focal Loss (\( L_{\text{AF}} \)).}  
This term places the strongest penalty on the cases where the classifier assigns a high predicted score \( p \) to an answer that is factually incorrect (i.e., labeled \( y = 0 \)). These are exactly the instances where MAD should be triggered but was mistakenly skipped. To address this, we adopt an asymmetric focal loss defined as:
\begin{equation}
    L_{\text{AF}}(y, p) := 
\begin{cases}
-\alpha_{1} (1 - p)^\gamma \log(p), & \text{if } y = 1,  \\
-\alpha_{0} ~p^\gamma \log(1 - p), & \text{if } y = 0,
\end{cases}
\end{equation}
where \( \gamma > 0 \) is a focusing parameter that down-weighs well-classified examples (i.e., when the predicted score \( p \) is close to the ground-truth label), and emphasizes incorrect or harder cases. The class-specific weights \( \alpha_1, \alpha_0 > 0 \) control the relative emphasis on each class. We typically set \( \alpha_0 > \alpha_1 \) to attribute a large penalty to cases where the model assigns a high \( p \) to incorrect cases (\( y = 0 \)), reflecting overconfident decisions that skip needed debate. This loss formulation directly addresses overconfidence by encouraging the classifier to assign a low predicted score \(p\) to incorrect single-agent answers. The asymmetric focal loss \( L_{\text{AF}} \) is thus designed to emphasize wrong predictions, many of which can be recovered through MAD, thus helping the classifier better identify cases that warrant debate.

\paragraph{Confidence Penalty (\( L_{\text{CP}} \)).} 
To further align the model's predicted score \( p \) with its uncertainty score \( u\), we introduce a regularization loss term that penalizes misalignment between them. This uncertainty score \( u\) reflects the model’s internal hesitation: a high value of \( u\) indicates distributed uncertainty across semantic interpretations, while a low value of \( u\) indicates peaked, confident predictions. Although \( p \) expresses the overall correctness belief, \( u \) offers a complementary perspective by capturing the uncertainty or hesitation embedded in the response features. To reconcile these two signals, we define Confidence Penalty as follows:
\begin{equation}
    L_{\text{CP}}(y, p, u) :=
    \begin{cases}
        u^2, & \text{if } y = 0 \text{ and } p > \tau, \\
        (1 - u)^2, & \text{if } y = 1 \text{ and } p < \tau, \\
        0, & \text{otherwise}.
    \end{cases}
\end{equation}
This confidence penalty term penalizes under-confidence for correct answers (high uncertainty occurs with a low $p$) and overconfidence for incorrect ones (low uncertainty occurs with a high $p$). By enforcing consistency between the predicted score $p$ and the uncertainty score $u$, this loss mitigates the misalignment between $p$ and the semantic hesitation signals, leading to more reliable debate triggering decisions.

\begin{table*}[t]
\centering
\setlength{\tabcolsep}{1.8pt}
\begin{tabular}{lcccccccccccc}
\toprule
 & \multicolumn{4}{c}{\textbf{Single-Agent}} & \multicolumn{4}{c}{\textbf{Full-Debate MAD}}& \multicolumn{4}{c}{\textbf{Selective MAD}} \\
\cmidrule(lr){2-5} \cmidrule(lr){6-9}\cmidrule(lr){10-13}

\multirow{1}{*}{\textbf{Dataset}} & \multicolumn{2}{c}{\textbf{CoT}} & \multicolumn{2}{c}{\textbf{SC}} & \multicolumn{2}{c}{\textbf{MAD}}  & \multicolumn{2}{c}{\textbf{\textsc{GD}}} & \multicolumn{2}{c}{\textbf{\textsc{DOWN}}} & \multicolumn{2}{c}{\textbf{iMAD}} \\
\cmidrule(lr){2-3} \cmidrule(lr){4-5} \cmidrule(lr){6-7}\cmidrule(lr){8-9}\cmidrule(lr){10-11}\cmidrule(lr){12-13}
& \textbf{Acc (\%)} & \textbf{\# Token} & \textbf{Acc (\%)} & \textbf{\# Token} & \textbf{Acc (\%)} & \textbf{\# Token} & \textbf{Acc (\%)} & \textbf{\# Token} & \textbf{Acc (\%)} & \textbf{\# Token} & \textbf{Acc (\%)} & \textbf{\# Token} \\
\midrule
MEDQA     & 76.6 & 653   & 77.3 & 3,482  & \underline{81.9} & 4,034  & 80.2 & 16,832 & 79.2 & 1,161 & \textbf{82.0} & 1,300 \\
MMLU      & 86.8 & 764   & 88.2 & 3,772  & \textbf{89.5} & 3,348  & 82.6 & 13,216 & 88.3 & 901 & \underline{89.2} & 1,010 \\
GSM8K     & 71.3 & 618   & 74.5 & 3,622  & \underline{76.4} & 3,446  & 73.4 & 15,321 & 72.6 & 812 & \textbf{84.8} & 1,025 \\ \midrule
OKVQA     & 88.3 & 1,945  & 89.2 & 11,031 & \underline{89.8} & 7,803 & 87.3 & 33,932 & 88.1 & 2,344 & \textbf{90.3} & 2,601 \\
VQA-v2       & 77.5 & 2,245  & 77.6 & 14,013 & 81.0 & 8,796 & \textbf{81.3} & 36,091 & 78.6 & 3,262 & \textbf{81.3} & 3,489 \\
ScienceQA & 86.0 & 1,720  & 86.2 & 9,833  & \underline{89.4} & 6,777 & 87.4 & 26,924 & 87.0 & 2,519 & \textbf{90.8} & 2,893 \\
\bottomrule
\end{tabular}
\caption{Accuracy (Acc) and average token cost (\# Token) per question comparison of \name and baselines across datasets. \textbf{Bold} values indicate the best result in each row, and \underline{underlined} values indicate the second best.}
\label{tab:fmad-performance}
\end{table*}

\paragraph{Expected Calibration Error (\( \text{ECE} \)).} 
To improve the reliability of \( p \) as a debate-triggering score, we incorporate ECE to encourage its alignment with empirical correctness. Suppose the dataset contains \( N \) instances indexed by \( i \in \{1, \dots, N\} \), where each instance $i$ has a predicted score \( p^{(i)} \in [0,1] \) and a binary ground-truth label \( y^{(i)} \in \{0,1\} \). Based on \( p^{(i)} \), We divide the interval \([0,1]\) into \( B \) equal-width bins and assign each instance $i$ to a bin \( b \in \{1, \dots, B\} \). Let \( \mathcal{I}_b \) denote the set of indices $i$ whose predicted scores fall into bin \( b \). We then compute ECE as:
\begin{equation}
\text{ECE} (y, p) := \sum_{b=1}^{B} \frac{1}{N} \left| \sum_{i \in \mathcal{I}_b} p^{(i)} -  \sum_{i \in \mathcal{I}_b} y^{(i)} \right|.
\end{equation}
For each bin, we measure the average absolute difference between predicted scores and ground-truth labels. Minimizing ECE aligns \(p\) with empirical correctness, leading to more reliable decision boundaries for triggering MAD.

%% file: 6_Evaluation.tex
\section{Evaluation}
\label{sec:evaluation}
We report token efficiency (i.e., token usage per question) and accuracy (i.e., final answer correctness) across six datasets. We also analyze \name's debate decisions on whether to trigger or skip MAD, and how often these decisions lead to beneficial outcomes. We further provide additional analysis of token efficiency and inference time in Appendix~\ref{sec:additional analysis}, and present additional results, including ablation studies on structured self-critique prompting and FocusCal loss, as well as cross-LLM evaluation, in Appendix~\ref{sec:additional results}.

\paragraph{Datasets.} We evaluate on six datasets across textual QA and image-text VQA. The QA benchmarks include (\romannumeral1) MedQA: USMLE-style medical QA~\cite{jin2020disease}; 
(\romannumeral2) MMLU: professional exam questions~\cite{hendrycks2020measuring}; 
and (\romannumeral3) GSM8K: grade-school math word problems~\cite{cobbe2021gsm8k}. 
For VQA, we use (\romannumeral4) OKVQA: emphasizes visual questions that require external knowledge~\cite{okvqa}; (\romannumeral5) VQA-v2: natural image QA with reduced bias~\cite{balanced_vqa_v2}; and (\romannumeral6) ScienceQA: science questions from school curricula~\cite{lu2022learn}. 

\paragraph{Baselines.} We compare \name with five strong baselines across single-agent, full-debate MAD, and selective MAD approaches. For single-agent methods, we include: (\romannumeral1) CoT~\cite{wei2022chain}; and (\romannumeral2) SC~\cite{wang2022self}, which runs CoT five times and selects the answer via majority voting. For full-debate MAD approaches that trigger debate for all instances, we consider: (\romannumeral3) MAD~\cite{liang-etal-2024-encouraging}, using agents with distinct personas in a three-agent setup; and (\romannumeral4) \textsc{GD}~\cite{liu2024groupdebateenhancingefficiencymultiagent}, which clusters 5 agents into subgroups for parallel discussion, followed by 3 rounds of inter-group consensus voting. For selective MAD triggering, we evaluate: 
(\romannumeral5) \textsc{DOWN}~\cite{eo2025debatenecessaryadaptivemultiagent}. Unlike other baselines, \textsc{DOWN} requires labeled evaluation data to tune its threshold. For fair comparison, we use a threshold of 0.8 as reported to be most effective in the original paper.

\paragraph{Metrics.} We evaluate each method using two key metrics: 
(\romannumeral1) total (input + output) token usage per data instance; (\romannumeral2) answer accuracy, measuring final answer correctness; (\romannumeral3) accuracy per 100k tokens (ApT), measuring token efficiency (see Appendix~\ref{sec:token-efficiency}); and (\romannumeral4) per-question inference time, measuring computational efficiency (see Appendix~\ref{sec:Inference Time Analysis}). 

\paragraph{Classifier Configurations.} 
We use a lightweight MLP with six fully connected layers of 200 hidden units each, batch normalization, ReLU activations, and a dropout rate of 0.2. We train the classifier for 50 epochs using the Adam optimizer~\cite{adam} with a learning rate of 0.001 on standardized features via StandardScaler~\cite{de2023choice}. To support task-agnostic generalization, we use PubMedQA~\cite{pubmedqa} and GQA~\cite{GQA} datasets for training, which are not included in the evaluation. We set the FocusCal loss hyperparameters as \( \alpha_{\text{0}} = 2.0 \), \( \alpha_{\text{1}} = 1.0 \), \( \gamma = 2 \), \( \lambda = 6 \), \( \mu = 5 \), and \( B = 15 \) bins for ECE. The debate threshold is set to \( \tau = 0.7 \). We chose these hyperparameters via grid search on a held-out validation set to ensure stable training and strong performance across all datasets.

\paragraph{Implementation Details.} For LLM-based prompting, we use Gemini 2.0 Flash as the LLM agents for the primary results and also use GPT-5 nano and Qwen 3.0 (see Appendix~\ref{sec:cross-llm} for details and results). By default, we use a temperature of 0.0 to ensure deterministic outputs and a maximum of 512 tokens. We present the detailed prompt templates in Appendix~\ref{sec:prompts}. We conducted MLP training and inference using a single NVIDIA RTX 4090 GPU.

\begin{table}[t]
\centering
\setlength{\tabcolsep}{2.0pt}
\small
\begin{tabular}{lcccccccc}
\toprule
\multirow{3}{*}{\textbf{Dataset}} 
& \multicolumn{4}{c}{\textbf{Skipped (\%)}}
& \multicolumn{4}{c}{\textbf{Triggered (\%)}} \\
\cmidrule(lr){2-5} \cmidrule(lr){6-9}
& \multicolumn{3}{c}{\textbf{Good}} 
& \textbf{Bad} 
& \multicolumn{3}{c}{\textbf{Bad}}
& \textbf{Good} \\
\cmidrule(lr){2-4} \cmidrule(lr){5-5} \cmidrule(lr){6-8} \cmidrule(lr){9-9}

& \ding{55}$\!\rightarrow$\ding{55} 
& \ding{51}$\!\rightarrow$\ding{51} 
& \ding{51}$\!\rightarrow$\ding{55} 
& \ding{55}$\!\rightarrow$\ding{51} 
& \ding{55}$\!\rightarrow$\ding{55} 
& \ding{51}$\!\rightarrow$\ding{51} 
& \ding{51}$\!\rightarrow$\ding{55}
& \ding{55}$\!\rightarrow$\ding{51} \\
\midrule
MEDQA     & 10.5 & 66.3 & 4.9 & 4.8  & 1.0 & 3.7 & 1.7 & 7.1 \\
MMLU      & 5.8  & 80.0 & 3.2 & 3.5 & 0.5 & 2.5 & 1.1  & 3.4 \\
GSM8K     & 6.8  & 57.1 & 11.3 & 2.9  & 2.7 & 0.3 & 2.7 & 16.2\\ \midrule
OKVQA     & 6.8  & 83.7 & 2.8 & 2.3  & 0.0 & 1.2 & 0.6 & 2.6\\
VQA-v2       & 12.9 & 69.7 & 5.3 & 4.9  & 0.4 & 2.1 & 0.4 & 4.3\\
ScienceQA & 6.1  & 78.0 & 3.6 & 2.2  & 0.0 & 3.5 & 0.9 & 5.7\\
\bottomrule
\end{tabular}
\caption{Breakdown of \name debate decisions by whether MAD was skipped or triggered, and whether the decision was beneficial (Good) or harmful (Bad).}
\label{tab:mad_decision_breakdown}
\end{table}


\subsection{Main Results} 
\label{sec:main-results}
We compare the token efficiency and accuracy of \name against single-agent, full-debate MAD, and selective MAD.

\paragraph{Advantage over Single-Agent and Full-Debate MAD.} As shown in Table~\ref{tab:fmad-performance}, \name achieves superior token efficiency while maintaining or improving accuracy over all single-agent and full-debate MAD baselines. While CoT uses the fewest tokens, \name achieves up to 13.5\% higher accuracy. Compared to SC, \name drastically reduces token usage with consistently higher accuracy. For example, on MEDQA, \name reduces token cost by 62.7\% while improving accuracy by 4.7\%. Against full-debate MAD baselines, \name reduces token usage significantly while achieving comparable or higher accuracy. On MEDQA, \name uses 68\% fewer tokens than MAD and 92\% fewer than \textsc{GD}, while achieving the highest accuracy. A key to \name's zero-shot efficiency is the classifier's strong generalization: we trained it on two representative datasets to capture diverse model reasoning behaviors, enabling it to learn generalizable model behaviors. \name's advantage over the full-debate strategy is especially evident on GSM8K, where \name outperforms MAD by 8.4\% in accuracy. The only exception is MMLU, where MAD performs slightly better. In Table~\ref{tab:mad_decision_breakdown}, the classifier skips debate in 3.5\% of questions where it would help and 3.4\% where a triggered debate fixes an error. This is because many MMLU questions are short and factual across diverse domains, wrong single-agent answers often sound fluent and confident, giving few hesitation cues and causing the classifier to miss some needed debates.

\paragraph{Advantage over Confidence-Based Selective MAD Triggering.}

As shown in Table~\ref{tab:fmad-performance}, \textsc{DOWN} and \name incur comparable token costs, with \textsc{DOWN} using slightly fewer tokens by omitting self-critique and skipping some needed debates. However, both \name and MAD consistently achieve higher accuracy than \textsc{DOWN}, since \textsc{DOWN} cannot tune its confidence thresholds in the zero-shot setting. In contrast, \name incurs slightly more tokens due to self-critique prompting and more necessary debates, but this cost is justified by improved identification of recoverable errors. For example, on OKVQA, \textsc{DOWN}'s accuracy remains near the single-agent baseline, revealing its inability to identify when debate is truly needed. This limitation arises because \textsc{DOWN} learns data-specific model behavior rather than generalizable behavior. In contrast, \name captures hesitation cues through a classifier trained on diverse features, enabling more accurate debate decisions and achieving higher accuracy with fewer tokens in zero-shot settings.

\subsection{Breakdown of \name Debate Decisions} \label{sec:breakdown}
To evaluate the effectiveness of {\name}’s selective debate triggering, we analyze its decision outcomes across all datasets in Table~\ref{tab:mad_decision_breakdown}. For each instance, we precompute the single-agent and MAD outputs to establish whether MAD improves the answer, serving as the ground truth for evaluating decisions. We then measure how often \name matches the beneficial outcomes. Overall, up to 95.9\% of \name's decisions are beneficial. When skipping debate, \name preserves correct answers ({\ding{51}}$\rightarrow${\ding{51}}) in 65-80\% of cases and avoids wasted computation on unrecoverable errors ({\ding{55}}$\rightarrow${\ding{55}}) by up to 13\%. When triggering debate, \name often recover incorrect answers ({\ding{55}}$\rightarrow${\ding{51}}). For example, \name successfully flips 16.2\% of cases on GSM8K and 7.1\% on MEDQA, approaching their respective upper bounds of 19.1\% and 11.9\% (see Table~\ref{tab:mad_flipping_stats}). Crucially, harmful decisions, such as overturning correct answers ({\ding{51}}$\rightarrow${\ding{55}}) or incurring unnecessary debate overhead ({\ding{55}}$\rightarrow${\ding{55}} and {\ding{51}}$\rightarrow${\ding{51}}), remain consistently low (around 5-10\%). These results highlight \name's ability to selectively trigger MAD only when it is likely to improve accuracy, while avoiding unnecessary token costs.

%% file: 8_Conclusion.tex
\section{Conclusion}
We presented \name, a token-efficient MAD framework that triggers debates only when it is likely to improve outcomes. It combines structured self-critique prompting with a lightweight debate decision classifier trained with FocusCal loss, enabling effective debate decisions based on single-agent responses without using evaluation data. Compared with five baselines, \name reduces token usage by up to 92\% while improving accuracy by up to 13.5\%. These results show that \name is a practical and scalable solution for collaborative reasoning in agentic LLM systems. Future work includes exploring adaptive or online learning approaches to reduce labeling costs during classifier training and further improve generalization, as discussed in Appendix~\ref{sec:discussion}.

%% file: 9_Acknowledgement.tex
\section*{Acknowledgments}
We sincerely thank the anonymous reviewers of the AAAI'26 for their constructive comments and suggestions. This research was supported in part by NSF grant CNS-2315851, the Commonwealth Cyber Initiative (CCI), and Virginia Tech Presidential Postdoctoral Fellowship.

%% file: 10_Appendix.tex
\clearpage
\appendix

\section*{Appendix}
\begin{appendices}
In the appendix, we include implementation details, additional performance analyses, additional evaluation results, and discussions. First, we provide the prompt templates and 41 features in our debate decision classifier in Appendix~\ref{sec:implementation details}. Then, we present additional performance analyses on token efficiency and inference time in Appendix~\ref{sec:additional analysis} and report ablation studies on the structured self-critique prompt and FocusCal loss, as well as cross-LLM results using Qwen 3.0 and GPT-5 nano in Appendix~\ref{sec:additional results}. Lastly, we discuss our assumptions, limitations, and future work in Appendix~\ref{sec:discussion}.

\section{Implementation Details}
\label{sec:implementation details}
We provide the exact prompt templates for all agents in QA and VQA tasks, including shared context, memory synchronization, role definitions and outputs, structured JSON formats, and placeholder fields in Appendix~\ref{sec:prompts}, and summarize all 41 interpretable features used in our debate decision classifier in Appendix~\ref{sec:features}.

\input{Prompt_and_Example}
\input{feature_table}
\input{features}

\section{Additional Analyses}
\label{sec:additional analysis}
In this section, we provide additional performance analyses. In Section~\ref{sec:token-efficiency}, we analyze accuracy per 100k total tokens (ApT) to show the token efficiency. In Section~\ref{sec:Inference Time Analysis}, we analyze per-question latency and explain how the mix of input and output tokens affects total inference time.

\input{accuracy_per_token}
\input{efficiency}

\section{Additional Evaluation Results}
\label{sec:additional results}
To further assess both the effectiveness of the two key design components in \name and its generality across different LLMs, we present two ablation studies and a cross-LLM evaluation. The ablation studies focus on (\romannumeral1) the structured self-critique prompting, which is designed to enable interpretable and informative feature extraction for downstream debate decision classification (Section~\ref{sec:Ablation Studies - Effectiveness of Structured Self-Critique Prompting}); and (\romannumeral2) the FocusCal loss, which is designed to guide the classifier skip debates that are unnecessary (when the single-agent answer is already correct or not recoverable by triggering MAD) and trigger debates when they are likely to correct an initially wrong answer (Section~\ref{sec:Ablation Studies - Effectiveness of FocusCal Loss}). Finally, we present the results of the cross-LLM evaluation in Section~\ref{sec:cross-llm}.
\input{Ablation}
\input{Cross-model}

\input{7_Discussion}

\end{appendices}

%% file: Prompt_and_Example.tex
\subsection{Prompt Templates}
\label{sec:prompts}

This section specifies the full prompt protocol used in our framework. We first introduce the agent roles in MAD, including how the Debater and Judge agents interact during the debate and what outputs are expected from each agent. We then describe how MAD maintains shared context and memory synchronization across these agents. In contrast, Debater agents are not required to follow any structured format and instead produce free-form natural language responses. We then introduce the prompt placeholders that are filled in at runtime with instance-specific values and the current debate state, so that each agent receives a prompt that reflects the correct question, option set, and transcript. Finally, we list the exact task-specific prompt templates for VQA and QA, that specify what \emph{input} each agent receives, the \emph{instruction} it follows, and the \emph{output} it generates.

\paragraph{Roles and Outputs.}
Prior to deciding whether to trigger MAD, \name first executes a self-critique stage with a single self-critique agent. This self-critique single agent produces a preliminary answer together with a step-by-step explanation and a mandatory self-critique. The debate decision classifier then takes this self-critique response as input and decides whether a debate is necessary. If MAD is triggered, the framework instantiates three distinct agent roles: the \emph{Debater (Affirmative)} agent, the \emph{Debater (Negative)} agent, and the \emph{Judge} agent. The Debater (Affirmative) agent takes the single-agent self-critique answer as its initial position and provides reasons and evidence to support this answer. In contrast, the Debater (Negative) agent reviews the affirmative reasoning and highlights potential issues. When the Debater (Negative) agent disagrees, it proposes an alternative option and provides its own justification. During each debate round, both debaters are required to select exactly one option from the candidate set and provide a concise rationale grounded in the given context (e.g., the visual content for VQA or the textual passage for QA). The Judge agent then assesses the arguments from both sides and produces a structured JSON output that tells the system whether to proceed to another debate round or to terminate the process and return the final answer.

\paragraph{Shared Context and Memory Synchronization.}
When MAD is triggered, at the beginning of each debate round, the system constructs a running transcript that includes the question, the initial CoT reasoning and self-critique, and all previous messages from every agent. This transcript is then inserted into the prompt of each agent, so they all see the same question, reasoning, and prior messages before generating their next response.
In other words, before each debate round, we synchronize the memory of all agents by giving them the same transcript. This memory synchronization ensures that agents always make decisions under a consistent context and keeps the multi-round discussion coherent.

\paragraph{Structured JSON Output.}
At the end of each debate round, the Judge agent generates a single, parsable JSON output containing four specific fields: \texttt{Preference}, \texttt{Supported Side}, \texttt{Reason}, and \texttt{Debate Answer}. The \texttt{Preference} field takes values ``\texttt{Yes}'' or ``\texttt{No}'' and controls whether another debate round is started. If \texttt{Preference} is set to ``\texttt{No}'' and the current number of rounds is still below a preset maximum number of debate rounds, the Judge is indicating that neither the affirmative nor the negative side has presented a sufficiently convincing argument to settle the dispute. Consequently, the system initiates the next round of debate to solicit further reasoning. We manually set this maximum number of debate rounds to a small constant to keep the token cost under control. In all our experiments, we set the debate at 5 rounds. If the maximum number of rounds is reached without the Judge ever setting \texttt{Preference} to ``\texttt{Yes}'', the system enters a finalization stage in which the Judge must select the best-supported answer from the existing arguments and produce a final decision. 

Conversely, if \texttt{Preference} is set to ``\texttt{Yes}'', the Judge has identified a clear superior argument from a preferred side, and the \texttt{Supported Side} field is then set to either ``\texttt{Affirmative}'' or ``\texttt{Negative}''. In this case, the debate terminates immediately. The system records the content of the \texttt{Debate Answer} field as the judge's current best answer selected from the option set, and the \texttt{Reason} field provides a brief natural language explanation of why this side and answer are preferred. For the finalization stage (e.g., after the maximum number of rounds is reached), the Judge produces a simplified JSON output containing only the \texttt{Reason} and the final \texttt{Debate Answer}. To make the control logic robust, we validate every JSON output returned by the Judge agent. We mark a JSON output as malformed whenever it has text outside the JSON object or misses required keys. We also reject outputs that use invalid field values (e.g., a \texttt{Preference} other than ``\texttt{Yes}'' or ``\texttt{No}'') or specify a \texttt{Debate Answer} that is not in the option set. In such cases, we discard the response and reissue the Judge prompt until a valid JSON output is obtained.

\paragraph{Placeholders.}
In our prompt templates, we use \emph{placeholders} as dynamic fields that are filled in at runtime to adapt the instructions to a specific question and debate round. Each placeholder is a symbolic token, such as \texttt{\{QUESTION\}} or \texttt{\{ROUND\}}, which will be replaced by the system with a concrete value before sending the prompt to an agent. For example, \texttt{\{QUESTION\}} is replaced by the actual question text, \texttt{\{OPTIONS\}} by the current multiple-choice option set, and \texttt{\{TRANSCRIPT\}} by the full history of messages in the current debate. All agents use these placeholders in their prompt templates, so each prompt reflects the current question, its multiple-choice option set, the up-to-date debate transcript, and other instance-specific information.

We group these placeholders into two categories: \emph{example fields} and \emph{debate state fields}. The \emph{example fields} describe the static information for a given instance. We use \texttt{\{QUESTION\}} for the question text, \texttt{\{OPTIONS\}} for the current multiple-choice option set of that question, \texttt{\{OPTIONS2\}} for a reduced option set constructed from \texttt{\{OPTIONS\}} by collecting all distinct options proposed by any agent during the debate and using this subset only in the final decision step, and \texttt{\{IMAGE\}} to refer to the visual input in VQA tasks. These fields do not change across debate rounds for the same example. \emph{The debate state fields} describe the dynamic status of the ongoing debate. We use \texttt{\{ROUND\}} to record the current round index and \texttt{\{TRANSCRIPT\}} to store the synchronized dialogue history, including the question, the initial reasoning or self-critique, and all previous messages from every agent. We further use \texttt{\{AFF\_ANS\}} and \texttt{\{NEG\_ANS\}} to denote the latest messages from the affirmative and negative debaters, respectively. From the perspective of the current speaker, \texttt{\{NEG\_ANS\}} refers to the opponent’s most recent message. 

Before each agent is prompted, the system first computes the current values of all relevant placeholders from the input example and the up-to-date transcript. The system then substitutes these values into the template. This placeholder substitution ensures that every agent receives a prompt that accurately reflects both the fixed information about the instance and the current state of the debate.

\paragraph{Prompt Template for QA Tasks.}
We use a different prompt template for each agent and execute them in a fixed order. We first apply the \emph{Self-Critique Single Agent} template. In this template, the agent receives \texttt{\{QUESTION\}} and \texttt{\{OPTIONS\}}, and is asked to choose an initial answer, explain it step by step, then deliberately argue against this answer, and finally state a final choice together with confidence scores for both the initial belief and the self-critique. The entire response is passed to the debate decision classifier. If the classifier decides to trigger MAD, we then use the two debater templates. The \emph{Debater (Affirmative)} template takes the synchronized transcript \texttt{\{TRANSCRIPT\}}, the question and option set, and the current affirmative stance \texttt{\{AFF\_ANS\}}. In the first debate round, it restates \texttt{\{AFF\_ANS\}} and provides supporting reasons. In later rounds, it also reads \texttt{\{NEG\_ANS\}} and either maintains its current answer or switches to a different option from \texttt{\{OPTIONS\}}, with a brief justification in each case. The \emph{Debater (Negative)} template also reads \texttt{\{TRANSCRIPT\}}, the question, and the option set. It examines the affirmative message \texttt{\{AFF\_ANS\}}, highlights potential issues, and when it disagrees, selects another option from \texttt{\{OPTIONS\}} and provides its own justification. Finally, we use two Judge templates. The per-round \emph{Judge} template asks the Judge agent to compare \texttt{\{AFF\_ANS\}} and \texttt{\{NEG\_ANS\}}, decide whether one side is clearly better, and return a structured JSON output, which the system uses to either start another round or stop the debate. The finalization Judge template is then used to list plausible candidates and commit to a single \texttt{Debate Answer} from \texttt{\{OPTIONS2\}} with a short reason to explain the decision.

\vspace{0.5em}
\begin{AIbox}{Self-Critique Single Agent}
\textbf{Input:} \texttt{\{QUESTION\}} and \texttt{\{OPTIONS\}}. \\
\textbf{Instruction:} Answer the multiple-choice question. Use a step-by-step chain of thought to justify your initial answer. Then, argue against your own answer with a plausible counter-reasoning — this is not optional. Finally, confirm your final answer and report confidence scores for both the initial belief and the forced disagreement. \\
\textbf{Output:} One option from (A/B/C/\dots), the initial reasoning with confidence score, and the self-critique reasoning with confidence score.
\end{AIbox}

\begin{AIbox}{Debater (Affirmative)}
\textbf{Input:} \texttt{\{TRANSCRIPT\}} (running log includes the initial single-agent answer, \texttt{\{AFF\_ANS\}} and \texttt{\{NEG\_ANS\}} choices with their arguments, and the Judge’s outputs from all previous debate rounds), \texttt{\{QUESTION\}}, \texttt{\{OPTIONS\}}, and \texttt{\{AFF\_ANS\}} (affirmative agent’s fixed initial stance, which is passed in every round to remind the agent of its side).\\
\textbf{Instruction:} First turn — restate \texttt{\{AFF\_ANS\}} and justify. Later turns — agree or refuse \texttt{\{NEG\_ANS\}}; if refusing, choose an alternative strictly from \texttt{\{OPTIONS\}} and justify. \\
\textbf{Output:} A short argument and one selected option from \texttt{\{OPTIONS\}}.
\end{AIbox}

\begin{AIbox}{Debater (Negative)}
\textbf{Input:} \texttt{\{TRANSCRIPT\}} (running log includes the initial single-agent answer, \texttt{\{AFF\_ANS\}} and \texttt{\{NEG\_ANS\}} choices with their arguments, and the Judge’s outputs from all previous debate rounds), \texttt{\{QUESTION\}}, \texttt{\{OPTIONS\}}, and latest affirmative message \texttt{\{AFF\_ANS\}} for the current round.\\
\textbf{Instruction:} Disagree when warranted. Provide an alternative strictly from \texttt{\{OPTIONS\}} and justify succinctly. \\
\textbf{Output:} A short counter-argument (or explicit agreement) and one selected option from \texttt{\{OPTIONS\}}.
\end{AIbox}

\begin{AIbox}{Judge (Per Round)}
\textbf{Input:} \texttt{\{TRANSCRIPT\}} (running log includes the initial single-agent answer, \texttt{\{AFF\_ANS\}} and \texttt{\{NEG\_ANS\}} choices with their arguments, and the Judge’s outputs from all previous debate rounds), \texttt{\{QUESTION\}}, \texttt{\{OPTIONS\}}, and \texttt{\{AFF\_ANS\}} and \texttt{\{NEG\_ANS\}} for the current round. \\
\textbf{Instruction:} Evaluate accuracy and justification quality; determine preference and tentative answer, if any. \\
\textbf{Output (JSON):} \{\texttt{Preference}: ``\texttt{Yes}'' $\vert$ ``\texttt{No}'', \texttt{Supported Side}: ``\texttt{Affirmative}'' $\vert$ ``\texttt{Negative}'', \texttt{Reason}: \dots, \texttt{Debate Answer}: one of \texttt{\{OPTIONS\}}\}.
\end{AIbox}

\begin{AIbox}{Judge (Finalization)}
\textbf{Input:} \texttt{\{TRANSCRIPT\}} (running log includes the initial single-agent answer, \texttt{\{AFF\_ANS\}} and \texttt{\{NEG\_ANS\}} choices with their arguments, and the Judge’s outputs from all previous debate rounds), \texttt{\{QUESTION\}}, \texttt{\{OPTIONS\}}, and reduced candidate list \texttt{\{OPTIONS2\}}. \\
\textbf{Instruction:} (A) List candidates strictly from \texttt{\{OPTIONS\}}. (B) Provide the final answer with a brief reason. \\
\textbf{Output (JSON):} \{\texttt{Reason}: \dots, \texttt{Debate Answer}:
one of \texttt{\{OPTIONS2\}}\}.
\end{AIbox}


\paragraph{Prompt Template for VQA Tasks.}
For VQA tasks, we reuse the same agent roles and control flow as in QA tasks, but extend every template to condition on the image and to explicitly mention visual evidence in the instructions. The \emph{Self-Critique Single Agent} template now takes \texttt{\{IMAGE\}}, \texttt{\{QUESTION\}}, and \texttt{\{OPTIONS\}} as input and begins with the instruction “From the image, answer the multiple-choice question,” so that both the initial answer and the self-critique are written with the image in view. The \emph{Debater (Affirmative)} and \emph{Debater (Negative)} templates also receive \texttt{\{IMAGE\}} together with the synchronized transcript, and their instructions explicitly ask the debaters to “justify from the image” or to “justify using visual evidence or logic” when agreeing, disagreeing, or proposing a new option. In practice, this encourages debaters to support their positions by referring to concrete visual cues in the image (e.g., objects, attributes, or spatial relations). Finally, the per-round and finalization \emph{Judge} templates keep the same JSON structure as in QA, but now take \texttt{\{IMAGE\}} as an additional input and explicitly instruct the Judge to “evaluate accuracy w.r.t. image and question” and to ensure that the preferred side and final \texttt{Debate Answer} are consistent with both the image and the debate dialogue history.

\vspace{0.5em}
\begin{AIbox}{Self-Critique Single Agent}
\textbf{Input:} \texttt{\{IMAGE\}}, \texttt{\{QUESTION\}}, and \texttt{\{OPTIONS\}}. \\
\textbf{Instruction:} From the image, answer the multiple-choice question. Use a step-by-step chain of thought to justify your initial answer. Then, argue against your own answer with a plausible counter-reasoning — this is not optional. Finally, confirm your final answer and report confidence scores for both the initial belief and the forced disagreement.\\
\textbf{Output:} One option (A/B/C/\dots), the initial reasoning with confidence score, and the self-critique reasoning with confidence score.
\end{AIbox}

\begin{AIbox}{Debater (Affirmative)}
\textbf{Input:} \texttt{\{TRANSCRIPT\}} (running log includes the initial single-agent answer, \texttt{\{AFF\_ANS\}} and \texttt{\{NEG\_ANS\}} choices with their arguments, and the Judge’s outputs from all previous debate rounds), \texttt{\{IMAGE\}}, \texttt{\{QUESTION\}}, \texttt{\{OPTIONS\}}, and \texttt{\{AFF\_ANS\}} (affirmative agent’s fixed initial stance, which is passed in every round to remind the agent of its side). \\
\textbf{Instruction:} First turn — restate \texttt{\{AFF\_ANS\}} and justify from the image. Later turns — agree or refuse \texttt{\{NEG\_ANS\}}; if refusing, select an alternative strictly from \texttt{\{OPTIONS\}} and justify briefly. \\
\textbf{Output:} A short argument and one selected option from \texttt{\{OPTIONS\}}.
\end{AIbox}

\begin{AIbox}{Debater (Negative)}
\textbf{Input:} \texttt{\{TRANSCRIPT\}} (running log includes the initial single-agent answer, \texttt{\{AFF\_ANS\}} and \texttt{\{NEG\_ANS\}} choices with their arguments, and the Judge’s outputs from all previous debate rounds), \texttt{\{IMAGE\}}, \texttt{\{QUESTION\}}, \texttt{\{OPTIONS\}}, and latest affirmative message \texttt{\{AFF\_ANS\}} for the current round. \\
\textbf{Instruction:} Disagree when warranted. Provide an alternative strictly from \texttt{\{OPTIONS\}} and justify using visual evidence or logic. \\
\textbf{Output:} A short counter-argument (or explicit agreement) and one selected option from \texttt{\{OPTIONS\}}.
\end{AIbox}

\begin{AIbox}{Judge (Per Round)}
\textbf{Input:} \texttt{\{TRANSCRIPT\}} (running log includes the initial single-agent answer, \texttt{\{AFF\_ANS\}} and \texttt{\{NEG\_ANS\}} choices with their arguments, and the Judge’s outputs from all previous debate rounds), \texttt{\{IMAGE\}}, \texttt{\{QUESTION\}}, \texttt{\{OPTIONS\}}, and \texttt{\{AFF\_ANS\}} and \texttt{\{NEG\_ANS\}} for the current round. \\
\textbf{Instruction:} Evaluate (1) accuracy w.r.t. image and question and (2) justification quality. Decide if there is a clear preference and provide a tentative answer if any. \\
\textbf{Output (JSON):} \{\texttt{Preference}: ``\texttt{Yes}'' $\vert$ ``\texttt{No}'', \texttt{Supported Side}: ``\texttt{Affirmative}'' $\vert$ ``\texttt{Negative}'', \texttt{Reason}: \dots, \texttt{Debate Answer}: one of \texttt{\{OPTIONS\}}\}.
\end{AIbox}

\begin{AIbox}{Judge (Finalization)}
\textbf{Input:} \texttt{\{TRANSCRIPT\}} (running log includes the initial single-agent answer, \texttt{\{AFF\_ANS\}} and \texttt{\{NEG\_ANS\}} choices with their arguments, and the Judge’s outputs from all previous debate rounds), \texttt{\{IMAGE\}}, \texttt{\{QUESTION\}}, \texttt{\{OPTIONS\}}, and reduced candidate list \texttt{\{OPTIONS2\}}. \\
\textbf{Instruction:} (A) List candidate answers strictly from \texttt{\{OPTIONS\}}. (B) Provide the final decision with a brief reason. \\
\textbf{Output (JSON):} \{\texttt{Reason}: \dots, \texttt{Debate Answer}: one of \texttt{\{OPTIONS2\}}\}.
\end{AIbox}

%% file: feature_table.tex
\begin{table*}[!t]
\centering
\setlength{\tabcolsep}{2.5pt}
\small
\begin{tabularx}{\textwidth}{lllX}
\toprule
\textbf{Source} & \textbf{Category} & \textbf{Feature Name} & \textbf{Description} \\
\midrule
\multirow{25}{*}{\shortstack[l]{\textbf{Input}\\\textbf{Question}}}
& \multirow{3}{*}{\shortstack[l]{Surface-level\\statistics}}
  & \texttt{QuestionToken} & \# of 
  tokens in the question \\
& & \texttt{AnswerToken} & \# of tokens in the model's final answer \\
& & \texttt{question\_named\_entity\_count} & \# of named entities (e.g., person, location, date) \\
\cmidrule{2-4}

& \multirow{11}{*}{\shortstack[l]{Uncertainty-related \\lexical cues}}
  & \texttt{qtype\_what} & Indicator if the question begins with “what” \\
& & \texttt{qtype\_where} & Indicator if the question begins with “where” \\
& & \texttt{qtype\_why} & Indicator if the question begins with “why” \\
& & \texttt{qtype\_how} & Indicator if the question begins with “how” \\
& & \texttt{qtype\_when} & Indicator if the question begins with “when” \\
& & \texttt{qtype\_who} & Indicator if the question begins with “who” \\
& & \texttt{qtype\_is} & Indicator if the question begins with “is” \\
& & \texttt{qtype\_are} & Indicator if the question begins with “are” \\
& & \texttt{qtype\_does} & Indicator if the question begins with “does” \\
& & \texttt{qtype\_do} & Indicator if the question begins with “do” \\
& & \texttt{qtype\_other} & Indicator for other question types \\
\cmidrule{2-4}
& Syntactic features
  & \texttt{QSyntacticDepth} & Maximum syntactic parse tree depth of the question \\
\cmidrule{2-4}
& \multirow{2}{*}{Readability metrics}
  & \texttt{Question\_flesch\_reading\_ease} & Flesch Reading Ease score of the question \\
& & \texttt{Question\_coleman\_liau\_index} & Coleman-Liau Index score of the question \\
\cmidrule{2-4}
& \multirow{3}{*}{\shortstack[l]{Part-of-speech\\counts}}
  & \texttt{Question\_num\_nouns} & \# of nouns in the question \\
& & \texttt{Question\_num\_verbs} & \# of verbs in the question \\
& & \texttt{Question\_num\_adjs} & \# of adjectives in the question \\

\midrule

\multirow{11}{*}{\shortstack[l]{\textbf{Initial}\\\textbf{Reasoning}}}
& \multirow{4}{*}{\shortstack[l]{Uncertainty-related \\lexical cues}}
  & \texttt{InitialConfidence} & Model’s confidence score for the initial answer \\
& & \texttt{InitialReason\_HedgeCount} & \# of hedge words (e.g., ``maybe'', ``likely'') \\
& & \texttt{InitialReason\_CertaintyCount} & \# of certainty words (e.g., ``definitely'', ``certainly'') \\
& & \texttt{InitialReason\_contrast} & \# of contrastive discourse markers (e.g., ``however'') \\
\cmidrule{2-4}
& Syntactic features
  & \texttt{InitialReason\_SyntacticDepth} & Max syntactic parse depth of the justification \\
\cmidrule{2-4}
& \multirow{2}{*}{Readability metrics}
  & \texttt{InitialReason\_flesch\_reading\_ease} & Flesch Reading Ease score of the justification \\
& & \texttt{InitialReason\_coleman\_liau\_index} & Coleman-Liau Index of the justification \\
\cmidrule{2-4}
& \multirow{3}{*}{\shortstack[l]{Part-of-speech\\counts}}
  & \texttt{InitialReason\_num\_nouns} & \# of nouns in the justification \\
& & \texttt{InitialReason\_num\_verbs} & \# of verbs in the justification \\
& & \texttt{InitialReason\_num\_adjs} & \# of adjectives in the justification \\
\midrule

\multirow{11}{*}{\textbf{\shortstack[l]{Self-\\Critique}}} 
& \multirow{5}{*}{\shortstack[l]{Uncertainty-related \\lexical cues}}
  & \texttt{NegativeConfidence} & Model’s confidence score for the alternative answer \\
& & \texttt{NegativeReason\_HedgeCount} & \# of hedge words in the critique \\
& & \texttt{NegativeReason\_CertaintyCount} & \# of certainty words in the critique \\
& & \texttt{NegativeReason\_contrast} & \# of contrastive discourse markers in the critique \\
& & \texttt{FinalConfidence} & Model’s confidence score for the final answer \\
\cmidrule{2-4}
& Syntactic features
  & \texttt{NegativeReason\_SyntacticDepth} & Max syntactic parse depth of the critique \\
\cmidrule{2-4}
& \multirow{2}{*}{Readability metrics}
  & \texttt{NegativeReason\_flesch\_reading\_ease} & Flesch Reading Ease score of the critique \\
& & \texttt{NegativeReason\_coleman\_liau\_index} & Coleman-Liau Index of the critique \\
\cmidrule{2-4}
& \multirow{3}{*}{\shortstack[l]{Part-of-speech\\counts}}
  & \texttt{NegativeReason\_num\_nouns} & \# of nouns in the critique \\
& & \texttt{NegativeReason\_num\_verbs} & \# of verbs in the critique \\
& & \texttt{NegativeReason\_num\_adjs} & \# of adjectives in the critique \\

\bottomrule
\end{tabularx}
\caption{List of extracted 41 features used in the debate decision classifier.}
\label{tab:feature_definitions}
\end{table*}

%% file: features.tex
\subsection{Features for Debate Decision Classification}
\label{sec:features}
In this section, we provide a detailed description of the features used by the debate decision classifier and evaluate their importance using SHAP and PCA. We then assess the necessity of features by removing the bottom 20\% of features based on a joint importance score that combines SHAP and PCA and comparing the accuracy and token costs with those of the full feature set.

\paragraph{Feature Details.}
To support lightweight and interpretable debate decisions, we extract 41 human-understandable linguistic and semantic features from three components of the structured single-agent response: the input question, the initial reasoning, and the self-critique. This feature set corresponds to Step~{\large \textcircled{\small 3}} \emph{Extracting Interpretable Features} in stage (ii) of our core design (see Section~\ref{sec:overview}). This design allows users to trace and understand why a debate is triggered or skipped. The debate decision classifier is based on a lightweight multi-layer perceptron (MLP) that operates on these features. It reuses the single-agent response and avoids extra LLM API calls or repeated sampling.

To provide more details on the feature set, we present all 41 extracted features and their descriptions in Table~\ref{tab:feature_definitions}, organized into five categories: (\romannumeral1) \emph{surface-level statistics (3 features from the input question only)}, such as token counts, which quantify question length and specificity, since very short or underdeveloped questions often correlate with lower reasoning quality; (\romannumeral2) \emph{readability metrics (6 features, consisting of 2 features for each of the input question, initial reasoning, and self-critique)}, including Flesch Reading Ease~\cite{davenport2014readabilitytweetsgeographiccorrelation} and the Coleman-Liau Index~\cite{coleman}, which assess how easy or complex a response is to read and potentially reveal vague or oversimplified reasoning; (\romannumeral3) \emph{syntactic features (3 features, consisting of one feature for each of the input question, initial reasoning, and self-critique)}, such as parse tree depth, which indicate grammatical complexity and the degree of elaboration in the reasoning structure; (\romannumeral4) \emph{part-of-speech counts (9 features, consisting of 3 features for each of the input question, initial reasoning, and self-critique}), including noun, verb, and adjective frequencies, which help characterize the factual density, action-oriented phrasing, and descriptive richness of the response; and (\romannumeral5) \emph{uncertainty-related lexical cues (20 features, consisting of 11 from the input question, 4 from the initial reasoning, and 5 from the self-critique)}, which reflect reasoning quality, such as hedging phrases (e.g., ``maybe''), certainty expressions (e.g., ``definitely''), and contrastive connectives (e.g., ``however'' and ``but''), and are commonly associated with hesitation or internal conflict in reasoning.

These features together capture fine-grained characteristics of the response that relate to model confidence, uncertainty, and reasoning behavior, enabling the classifier to determine when MAD is likely to be beneficial based on the single-agent output. Since all features are derived directly from this structured output without requiring additional queries or repeated sampling, the classifier can make efficient and accurate decisions while preserving scalability and improving token efficiency.

\begin{figure*}[t]
\centering
\begin{subfigure}[t]{0.9\linewidth}
\centering
\includegraphics[width=\linewidth]{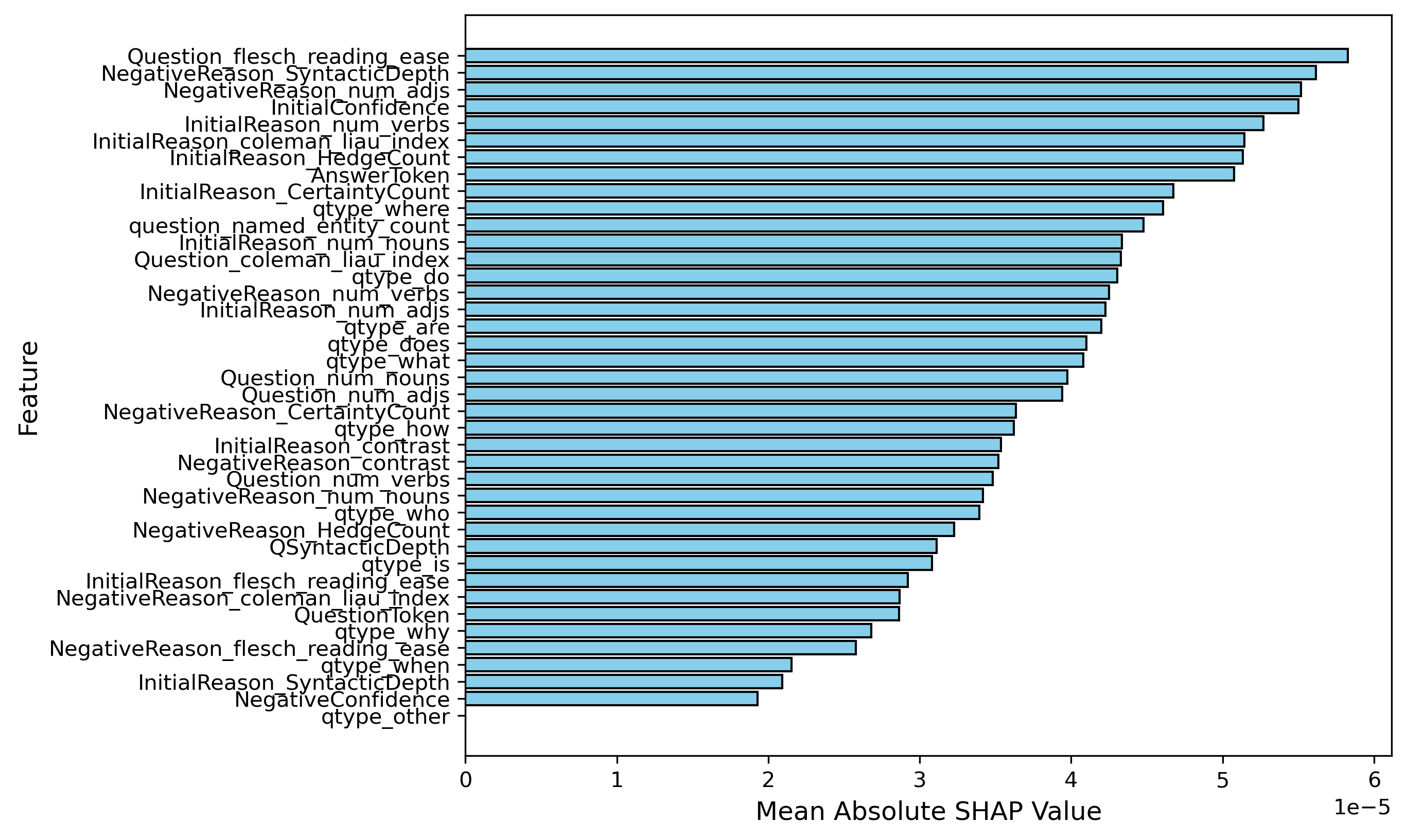}
\caption{SHAP Importance}
\label{fig:shap-importance}
\end{subfigure}
\hfill
\begin{subfigure}[t]{0.9\linewidth}
\centering
\includegraphics[width=\linewidth]{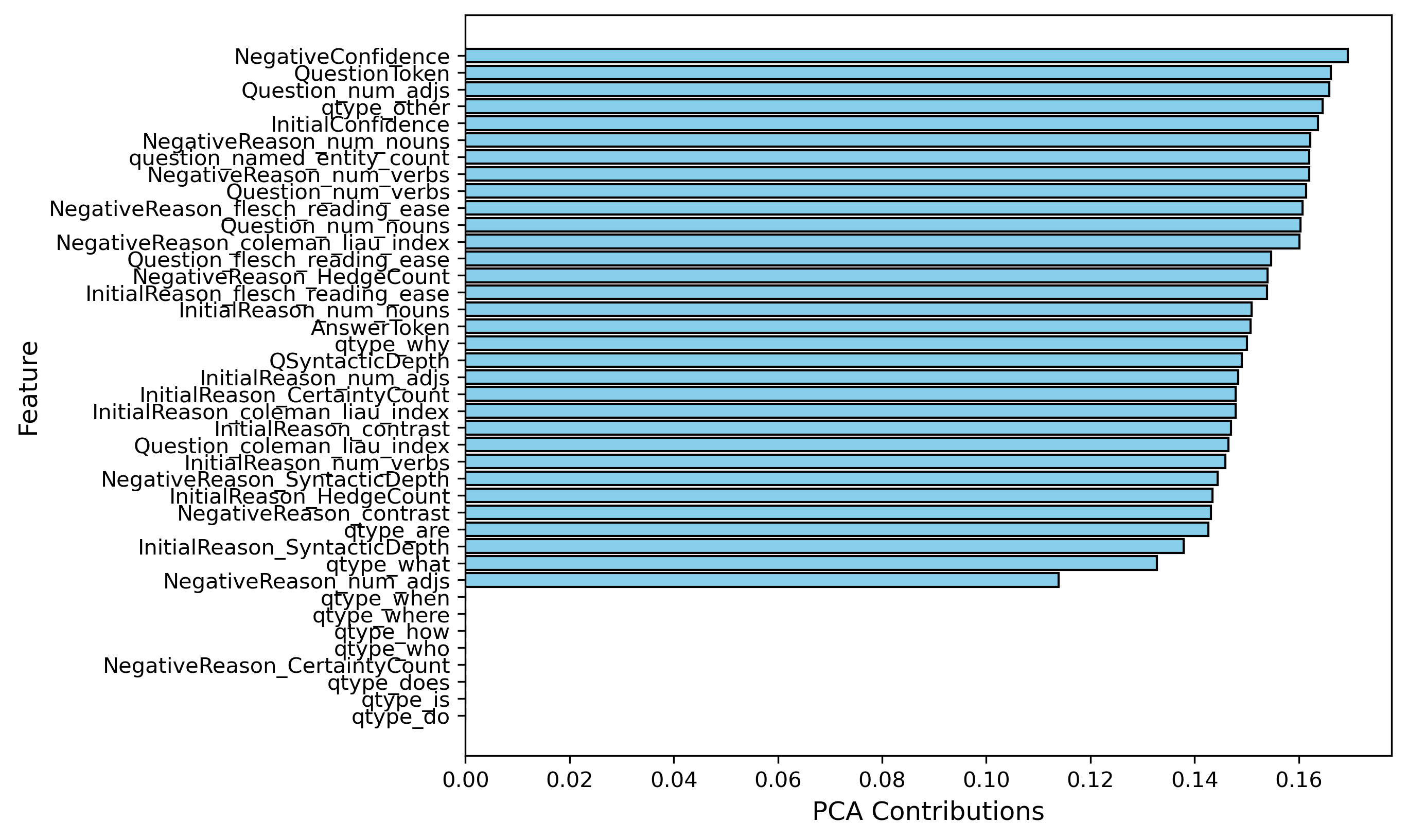}
\caption{Aggregated PCA Contributions}
\label{fig:pca-importance}
\end{subfigure}
\caption{The feature importance of 41 features used by the debate decision classifier: (a) mean absolute SHAP values and (b) aggregated PCA contribution for each feature, showing how much that feature contributes to the principal components that explain most of the variance across the dataset.}
\label{fig:feature-importance}
\end{figure*}

\paragraph{Feature Importance with SHAP and PCA.}
To understand which features drive the debate decision, we evaluate feature importance using two complementary metrics over all 41 features. First, we use SHapley Additive exPlanations (SHAP) importance~\cite{SHAP}, a feature-attribution method that has been widely used in the NLP and LLM settings to interpret model behavior~\cite{horovicz-goldshmidt-2024-tokenshap, SHAP-LLM-2024}. For each feature, SHAP importance measures how much that feature contributes to the classifier’s predicted debate-triggering score on each data instance. For each feature, we compute the mean absolute SHAP value (typically ranging from $1.9\times10^{-5}$ to $5.8\times10^{-5}$ in our setting) and report these values in Fig.~\ref{fig:shap-importance}, where longer bars indicate stronger influence on the classifier’s decisions. Second, contributions from Principal Component Analysis (PCA)~\cite{pca} summarize how much each feature influences the top principal components, which are the directions in which the feature values vary the most across examples. Concretely, we run PCA on the standardized feature matrix, sum the absolute loadings of each feature over the principal components that explain most of the variance, and plot these aggregated contributions (typically ranging from 0.02 to 0.15 per feature) in Fig.~\ref{fig:pca-importance}. Similar to SHAP, longer bars in PCA features account for higher variance in the feature space.

Across both metrics, features derived from the self-critique consistently rank among the most influential. In particular, features capturing uncertainty and conflict signals (e.g., \texttt{NegativeReason HedgeCount}, \texttt{NegativeReason contrast}, \texttt{NegativeReason SyntacticDepth}, and readability features) appear near the top in both SHAP importance and PCA contributions. In addition, several question-level features such as \texttt{QuestionToken} and the question-type indicators (e.g., \texttt{qtype why} and \texttt{qtype how}) also contribute noticeably. Intuitively, longer questions and open-ended types like ``why'' or ``how'' often correspond to harder problems, so combining these signals with the uncertainty patterns in the self-critique helps the classifier better identify cases where debate is likely to be beneficial. Overall, the SHAP and PCA results indicate that the classifier mainly relies on self-critique features. It also uses readability, uncertainty-related lexical cues, and syntactic features from the question and initial reasoning to decide when triggering debate is likely to be beneficial.

\paragraph{Feature Necessity.}
To further investigate whether low-ranked features are still useful, we examine the impact of these features on the classifier. For each feature, we first compute a joint importance score by normalizing its absolute SHAP value and its PCA contribution and then summing these two normalized quantities. The SHAP term reflects how strongly the feature affects the classifier’s prediction on individual examples, while the PCA term reflects how much the feature contributes to variation across the entire dataset. We then sort all 41 features by this joint score and remove the bottom 20\% of features. Specifically, we remove the following eight features that fall into the bottom 20\% by joint score: \texttt{qtype\_do}, \texttt{qtype\_who}, \texttt{qtype\_when}, \texttt{qtype\_how}, \texttt{qtype\_where}, \texttt{qtype\_is}, \texttt{qtype\_does}, and \texttt{NegativeReason\_CertaintyCount}. We retrain the debate decision classifier using the remaining 33 features and compare it with the full model that uses all 41 features. For both versions, we report accuracy and the average token costs per question on each dataset. 

As shown in Table~\ref{tab:feature_prune}, removing these low-ranked features results in a small but consistent accuracy drop of about 0.5\% on average and a noticeable increase in token usage of about 6.9\% across the six datasets (from 2,053 to 2,194 tokens per question). The higher token cost arises since the pruned classifier makes less precise decisions about when debate is needed, triggering debate more often on questions that are already answered correctly. The accuracy drop occurs since the pruned classifier also misses some hard questions where debate could have corrected the initial answer. These results indicate that even features with low SHAP and PCA scores still help the classifier make more accurate and token-efficient decisions. Therefore, we keep all 41 features in our final design.

\begin{table}[t]
\centering
\setlength{\tabcolsep}{5.5pt}
\begin{tabular}{lcccc}
\toprule
\multirow{2}{*}{\textbf{Dataset}} &
\multicolumn{2}{c}{\textbf{\shortstack[c]{\name\\(w/o Bottom 20\%)}}} &
\multicolumn{2}{c}{\textbf{\shortstack[c]{\name\\(All Features)}}} \\
\cmidrule(lr){2-3}\cmidrule(lr){4-5}
& \textbf{Acc (\%)} & \textbf{\# Token} & \textbf{Acc (\%)} & \textbf{\# Token} \\
\midrule
MEDQA     & 81.7 & 1,381 & 82.0 & 1,300 \\
MMLU      & 88.7 & 1,061 & 89.2 & 1,010 \\
GSM8K     & 84.0 & 1,107 & 84.8 & 1,025 \\\midrule
OKVQA     & 90.0 & 2,783 & 90.3 & 2,601 \\
VQA-v2       & 80.4 & 3,633 & 81.3 & 3,489 \\
ScienceQA & 90.8 & 3,196 & 90.8 & 2,893 \\\midrule
Average   & 85.9 & 2,194 & 86.4 & 2,053 \\
\bottomrule
\end{tabular}
\caption{Feature necessity study in \name. We normalize the SHAP importance scores and PCA loadings for each feature, sum them to obtain a joint importance score, and remove the bottom 20\% of features. It shows an accuracy drop of about 0.5\% with the higher token usage of about 6.9\% per question on average, showing the importance of all 41 features.}
\label{tab:feature_prune}
\end{table}

%% file: accuracy_per_token.tex
\begin{table*}[t]
\centering
\setlength{\tabcolsep}{4.5pt} 
\begin{tabular}{@{}lcccccccccccc@{}}
\toprule
\multirow{4}{*}{\textbf{Dataset}} 
& \multicolumn{4}{c}{\textbf{Single-Agent}} 
& \multicolumn{4}{c}{\textbf{Full-Debate MAD}} 
& \multicolumn{4}{c}{\textbf{Selective MAD}} \\
\cmidrule(lr){2-5} \cmidrule(lr){6-9} \cmidrule(lr){10-13}
& \multicolumn{2}{c}{\textbf{CoT}} 
& \multicolumn{2}{c}{\textbf{SC}} 
& \multicolumn{2}{c}{\textbf{MAD}} 
& \multicolumn{2}{c}{\textbf{\textsc{GD}}} 
& \multicolumn{2}{c}{\textbf{\textsc{DOWN}}} 
& \multicolumn{2}{c}{\textbf{\name}} \\
\cmidrule(lr){2-3} \cmidrule(lr){4-5} \cmidrule(lr){6-7} \cmidrule(lr){8-9} \cmidrule(lr){10-11} \cmidrule(lr){12-13}
& \textbf{ApT} & \textbf{Acc. (\%)} 
& \textbf{ApT} & \textbf{Acc. (\%)} 
& \textbf{ApT} & \textbf{Acc. (\%)} 
& \textbf{ApT} & \textbf{Acc. (\%)} 
& \textbf{ApT} & \textbf{Acc. (\%)} 
& \textbf{ApT} & \textbf{Acc. (\%)} \\
\midrule
MEDQA     
& 117.3 & 76.6 & 22.2 & 77.3 
& 20.3 & 81.9 & 4.8  & 80.2 
& 68.2 & 79.2 & 63.1 & 82.0 \\
MMLU      
& 113.6 & 86.8 & 23.4 & 88.2 
& 26.7 & 89.5 & 6.2  & 82.6 
& 98.0 & 88.3 & 88.3 & 89.2 \\
GSM8K     
& 115.4 & 71.3 & 20.6 & 74.5 
& 22.2 & 76.4 & 4.8  & 73.4 
& 89.4 & 72.6 & 82.7 & 84.8 \\
\midrule
OKVQA     
& 45.4  & 88.3 & 8.1  & 89.2    
& 11.5  & 89.8 & 2.6  & 87.3  
& 37.6  & 88.1 & 34.7 & 90.3 \\
VQA-v2    
& 34.5  & 77.5 & 5.5  & 77.6    
& 9.2   & 81.0 & 2.3  & 81.3  
& 24.1  & 78.6 & 23.3 & 81.3 \\
ScienceQA 
& 50.0  & 86.0 & 8.8  & 86.2    
& 13.2  & 89.4 & 3.2  & 87.4  
& 34.5  & 87.0 & 31.4 & 90.8 \\
\midrule
Average   
& 79.4  & 81.1 & 14.8 & 82.2   
& 17.2  & 84.7 & 4.0  & 82.0  
& 58.6  & 82.3 & 53.9 & 86.4 \\
\bottomrule
\end{tabular}
\caption{Accuracy per 100k tokens (ApT) and corresponding accuracy (Acc.) from Table~\ref{tab:fmad-performance}. ApT measures the expected number of correct answers obtained for every 100k tokens, so larger values indicate better token efficiency.}
\label{tab:apt}
\end{table*}

\subsection{Analysis on Accuracy per Token (ApT)}
\label{sec:token-efficiency}

To further evaluate token efficiency, we report \emph{Accuracy per Tokens} (ApT) as an auxiliary metric, following recent works that measure token efficiency in terms of ApT for reasoning LLMs~\cite{yan-etal-2025-long, bi2025cotxadaptiveframeworkcrossmodel}. For each method and dataset, we measure the average total number of tokens (input + output) used per question. Given accuracy \(\text{Acc} \in [0,1]\) and the average total number of tokens (i.e., \(\#~\text{Token}\) in Table~\ref{tab:fmad-performance}), ApT is defined as:
\[
\mathrm{ApT} = \text{Acc} \times \frac{100,000}{\#~\text{Token}}.
\]
A larger ApT means that, on average, a method produces more correct answers per 100,000 tokens, but ApT does not capture the final accuracy level. Some methods that use very few tokens (e.g., CoT) can appear highly efficient under ApT while still performing much worse in accuracy than multi-agent methods. Therefore, ApT should be viewed only as a coarse indicator of token usage and must be interpreted together with accuracy. In practice, a method with slightly lower ApT but substantially higher accuracy is usually more desirable than a method that maximizes ApT alone.

In Table~\ref{tab:apt}, we report ApT and the corresponding accuracy across all datasets. CoT attains the highest ApT as it uses very few tokens per question, but its accuracy is still lower than that of multi-agent methods. SC improves accuracy over CoT by running multiple CoT generations and aggregating the outputs, but these extra generations substantially increase the token cost. As a result, SC’s ApT becomes much smaller, since token usage grows faster than the accuracy gain. Full-debate MAD baselines such as MAD and \textsc{GD} achieve higher accuracy than single-agent methods but need a larger token cost, since they trigger debate for every question and repeatedly broadcast the growing transcript to all agents. This high cost leads to low ApT values. 

Selective debate methods (e.g., \textsc{DOWN} and \name) strike a better balance between accuracy and token usage. Both \textsc{DOWN} and \name obtain much higher ApT than MAD and \textsc{GD} since they skip debating on many questions that the single agent already answers correctly. Compared with \textsc{DOWN}, \name has slightly lower ApT on average (53.9 vs. 58.6) across the six datasets. This is because \name still triggers debates on questions where debate is necessary to improve the answer, while \textsc{DOWN} skips some of these cases. These additional debates consume extra tokens, so ApT becomes slightly lower even though \name achieves higher accuracy. 
\name can be applied in a zero-shot setting, where it uses a fixed debate decision classifier without any dataset-specific tuning, while \textsc{DOWN} requires access to labeled evaluation data to tune its confidence threshold, violating the standard zero-shot assumption typically upheld by single-agent and MAD baselines. Under this stricter zero-shot protocol for \name, the modest extra token cost still yields an average accuracy gain of about 4.1\% over \textsc{DOWN}.

In practice, this trade-off is often favorable: \name retains most of the token savings of selective debate while achieving higher accuracy consistently across datasets. ApT shows that \name preserves strong token efficiency compared with full-debate MAD and single-agent baselines, while the accuracy results highlight its consistent gains over \textsc{DOWN}.

%% file: efficiency.tex
\begin{table}[t]
\centering
\label{tab:efficiency-performance}
\setlength{\tabcolsep}{2.8pt}
\begin{tabular}{lcccccc}
\toprule
 \multirow{3.5}{*}{\textbf{Dataset}}& \multicolumn{2}{c}{{\textbf{Single-Agent}}} & \multicolumn{2}{c}{{\textbf{Full-Debate}}}& \multicolumn{2}{c}{\textbf{Selective MAD}}  \\
 & \multicolumn{2}{c}{{\textbf{(s)}}} & \multicolumn{2}{c}{\textbf{MAD (s)}}& \multicolumn{2}{c}{\textbf{(s)}}  \\
\cmidrule(lr){2-3} \cmidrule(lr){4-5}\cmidrule(lr){6-7} 

 & {\textbf{CoT}} &{\textbf{SC}} & {\textbf{MAD}}  & {\textbf{\textsc{GD}}} & {\textbf{\textsc{DOWN}}} & {\textbf{\name}} \\
\midrule
MEDQA     & 1.4 & 9.4   & 11.5 & 49.2  & 1.8 & 1.8  \\
MMLU      & 1.0 & 5.3  & 5.1 & 29.7  & 1.1 & 1.1  \\
GSM8K     & 1.6 & 8.7   & 8.0 & 29.8  & 1.6 & 1.6   \\ \midrule
OKVQA     & 1.4 & 12.9  & 9.1 & 37.1 & 1.7 & 1.7  \\
VQA-v2       & 1.3 & 14.9 & 8.1 & 33.6 & 1.5 & 1.6  \\
ScienceQA & 1.5 & 15.2  & 10.9 & 44.3  & 1.8 & 1.8  \\
\bottomrule
\end{tabular}
\caption{Inference time per question comparison of iMAD and baselines across datasets.
}
\label{tab:inference-time}
\end{table}

\begin{table*}[t]
\centering
\begin{tabular}{lcccccccccccc}
\toprule
& \multicolumn{4}{c}{\textbf{Single-Agent}} 
& \multicolumn{4}{c}{\textbf{Full-Debate MAD}} 
& \multicolumn{4}{c}{\textbf{Selective MAD}} \\
\cmidrule(lr){2-5} \cmidrule(lr){6-9} \cmidrule(lr){10-13}
\multicolumn{1}{c}{\textbf{Dataset}} 
& \multicolumn{2}{c}{\textbf{CoT}} 
& \multicolumn{2}{c}{\textbf{SC}} 
& \multicolumn{2}{c}{\textbf{MAD}}  
& \multicolumn{2}{c}{\textbf{\textsc{GD}}} 
& \multicolumn{2}{c}{\textbf{\textsc{DOWN}}} 
& \multicolumn{2}{c}{\textbf{iMAD}} \\
\cmidrule(lr){2-3} \cmidrule(lr){4-5} \cmidrule(lr){6-7} \cmidrule(lr){8-9} \cmidrule(lr){10-11} \cmidrule(lr){12-13}
& \textbf{In} & \textbf{Out} 
& \textbf{In} & \textbf{Out} 
& \textbf{In} & \textbf{Out} 
& \textbf{In} & \textbf{Out} 
& \textbf{In} & \textbf{Out} 
& \textbf{In} & \textbf{Out} \\
\midrule
MEDQA      
&   478 & 175 &  2,307 & 1,175 &  2,834 & 1,200 & 10,682 & 6,150 &   981 & 180 & 1,110 & 190 \\
MMLU       
&   610 & 154 &  2,956 &   816 &  2,565 &   783 &  8,642 & 4,574 &   732 & 169 &   840 & 170 \\
GSM8K      
&   415 & 203 &  2,518 & 1,104 &  2,635 &   811 & 11,542 & 3,779 &   609 & 203 &   822 & 203 \\
\midrule
OKVQA      
& 1,865 &  80 & 10,294 &   737 &  7,463 &   340 & 31,812 & 2,120 & 2,154 & 190 & 2,396 & 205 \\
VQA-v2     
& 2,175 &  70 & 13,211 &   802 &  8,346 &   450 & 34,282 & 1,809 & 3,077 & 185 & 3,291 & 198 \\
ScienceQA  
& 1,635 &  85 &  8,972 &   861 &  6,290 &   487 & 24,414 & 2,510 & 2,329 & 190 & 2,690 & 203 \\
\bottomrule
\end{tabular}
\caption{The number of Input (In) tokens and output (Out) tokens per question comparison of iMAD and baselines across datasets. The sum of input token and output token equals the total \# Token reported in Table~\ref{tab:fmad-performance}.}
\label{tab:io-token-decomp-full}
\end{table*}

\subsection{Inference Time Analysis}
\label{sec:Inference Time Analysis}
To evaluate the computational efficiency of \name, we measure the per-question inference time for each method on each dataset. Here, inference time refers to the time spent to process a single question and produce an answer, including all calls to the underlying LLM.
As shown in Table~\ref{tab:inference-time}, CoT achieves the lowest inference time but yields lower accuracy compared with multi-agent methods. SC improves accuracy over CoT by running CoT multiple times and aggregating the outputs. However, these extra generations substantially increase the overall computation, leading to longer inference times (up to 5.3--15.2~s per question). Full-debate MAD frameworks such as MAD and \textsc{GD} have much longer inference time, reaching up to 49.2~s per question, since they trigger debate for every question and repeatedly resend a growing shared transcript to all agents. In contrast, selective MAD methods such as \textsc{DOWN} and \name keep the inference time close to that of a single agent, ranging from 1.1~s to 1.8~s per question. Unlike \textsc{DOWN}, which often achieves lower accuracy due to its limited ability to generalize across different models or datasets, \name consistently matches or exceeds the accuracy of full-debate MAD frameworks while keeping inference time close to single-agent baselines.

On average, \name reduces inference time by 32.6\% compared with MAD and by 68.9\% compared with \textsc{GD}, while still maintaining low per-question inference time. It is important to note that inference time is not directly proportional to the total number of tokens. Table~\ref{tab:fmad-performance} reports the total number of tokens (input and output), and Table~\ref{tab:io-token-decomp-full} decomposes them into input and output tokens. While additional input tokens incur a slightly longer processing time, generating output tokens is typically more time-consuming since the model needs to generate them sequentially. For example, GSM8K uses fewer total tokens than MMLU (618 vs. 764 tokens), yet produces more output tokens on average (203 vs. 154 tokens). As a result, GSM8K has a longer inference time (1.6~s vs. 1.0~s). However, a very large number of input tokens can also dominate the inference time. For example, within the VQA-v2 dataset, SC uses fewer output tokens than MAD (802 vs. 2,450 tokens) but a much larger number of input tokens (13,211 vs. 6,346 tokens) as it aggregates multiple single-agent runs. This much larger number of input tokens is the reason why SC has a longer inference time than MAD despite generating fewer output tokens.

Overall, these findings highlight that \name provides a favorable balance between accuracy and computational cost. It achieves accuracy comparable to or better than full-debate MAD methods while keeping inference time close to that of single-agent baselines, making it a practical choice for real-world deployments where both accuracy and inference speed are critical.

%% file: Ablation.tex
\subsection{Ablation Study on Structured Self-Critique Prompt Design}
\label{sec:Ablation Studies - Effectiveness of Structured Self-Critique Prompting}
To evaluate the impact of prompt design, we compare the structured self-critique prompt used in \name with a standard CoT prompt that does not include self-critique. In the structured prompt, the model is asked to (\romannumeral1) provide an initial answer with step-by-step reasoning; and (\romannumeral2) explicitly critique its own answer before committing to a final choice. In the standard CoT prompt, the model only produces step-by-step reasoning for a single answer without any self-critique. For this study, we report two quantities for each method and each dataset: (\romannumeral1) total (input + output) token usage per data instance, and (\romannumeral2) accuracy of the answer (i.e., final answer correctness). These two metrics allow us to see how the self-critique prompt affects both token usage and accuracy performance. As shown in Table~\ref{tab:cot_vs_fmad_prompt}, the structured self-critique prompt in \name consistently improves accuracy across all datasets and tasks compared with the standard CoT prompt. The gain is more evident for complex reasoning tasks. For example, on math reasoning benchmarks such as GSM8K, the self-critique prompt improves accuracy by 7.2\% over the standard CoT prompt. Although the structured self-critique prompt slightly increases token usage (e.g., from 1,222 to 1,300 tokens per question on MEDQA), this extra cost is small compared with the accuracy gain and is thus desirable in practice.

This accuracy improvement comes from the ability of the structured self-critique prompt to elicit more detailed and diverse reasoning. The prompt asks the model not only to justify its initial answer but also to generate counterarguments that challenge it. For example, an initial reasoning may say ``The answer is option A because \dots,'' while the self-critique then says ``However, option B could also be correct because \dots'' or ``I am less certain about A after considering B.'' These additional steps often reveal signals that are missing from standard CoT outputs, such as explicit mentions of alternative options, phrases that indicate doubt (e.g., ``maybe'', ``not sure'', and ``could be''), or changes of mind between the initial answer and the final answer. 

These signals help the debate decision classifier decide whether a debate is needed. 
For instance, if the self-critique proposes a new option and relies heavily on hedging phrases while comparing it with the original answer, the classifier can interpret this pattern as evidence of uncertainty, thereby triggering MAD. In contrast, if the self-critique repeats the same answer, gives a consistent justification, and contains few hedging words, the classifier is more likely to skip the debate. Standard CoT prompts usually produce only a single chain of reasoning that supports one answer, which makes it hard to distinguish questions that really need debate from those that do not.

\begin{table}[t]
\centering
\setlength{\tabcolsep}{5.5pt}
\begin{tabular}{lcccc}
\toprule
\multirow{1}{*}{\textbf{Dataset}}
& \multicolumn{2}{c}{\shortstack{\textbf{Prompt w/o} \\ \textbf{Self-Critique}}}
& \multicolumn{2}{c}{\shortstack{\textbf{Prompt w/} \\ \textbf{Self-Critique}}}\\

\cmidrule(lr){2-3} \cmidrule(lr){4-5}
& \textbf{Acc (\%)} & \textbf{\# Token}
& \textbf{Acc (\%)} & \textbf{\# Token}  \\
\midrule
MEDQA     & 80.2 & 1,222 & 82.0 & 1,300 \\
MMLU      & 88.3 & 943 & 89.2 & 1,010 \\
GSM8K     & 77.6 & 988 & 84.8 & 1,025 \\ \midrule
OKVQA     & 88.1 & 2,342 & 90.3 & 2,601 \\
VQA-v2       & 78.9 & 3,314 & 81.3 & 3,489 \\
ScienceQA & 88.4 & 2,541 & 90.8 & 2,893 \\
\bottomrule
\end{tabular}
\caption{Performance of single-agent prompts with and without self-critique across QA and VQA datasets.}
\label{tab:cot_vs_fmad_prompt}
\end{table}


\subsection{Ablation Study on FocusCal Loss}
\label{sec:Ablation Studies - Effectiveness of FocusCal Loss}
To evaluate the effectiveness of the proposed FocusCal loss in \name, we conduct ablation studies from three perspectives: (\romannumeral1) how each loss component (i.e., $L_{\text{AF}}$, $L_{\text{CP}}$, and ECE) in FocusCal contributes to the final performance in terms of accuracy and token costs; (\romannumeral2) how FocusCal outperforms standard loss designs, such as Binary Cross-Entropy (BCE)~\cite{li2024rediscoveringbcelossuniform} and Mean Squared Error (MSE)~\cite{koksoy2006multiresponse}, on debate decision correctness;
and (\romannumeral3) what is the impact of the calibration term ECE in FocusCal by replacing ECE with BCE or MSE while keeping other terms of $L_{\text{AF}}$ and $L_{\text{CP}}$ fixed.

\begin{table}[t]
\centering
\begin{tabular}{lcc}
\toprule
\textbf{Loss Component} & \textbf{Acc (\%)} & \textbf{\# Token} \\
\midrule
$L_{\text{AF}}$ only               & 78.8 & 3,558 \\
$L_{\text{CP}}$ only               & 78.1 & 3,379 \\
ECE only                            & 79.1 & 3,757 \\ \midrule
$L_{\text{AF}}$ + $L_{\text{CP}}$  & 79.7 & 3,552 \\
$L_{\text{AF}}$ + ECE              & 79.7 & 3,561 \\
$L_{\text{CP}}$ + ECE              & 79.8 & 3,577 \\ \midrule
$L_{\text{AF}}$ + $L_{\text{CP}}$ + ECE (\textbf{FocusCal}) & \textbf{81.3} & \textbf{3,489} \\
\bottomrule
\end{tabular}
\caption{Ablation study on three loss terms in FocusCal using the VQA-v2 dataset in terms of accuracy (Acc) and total token cost (\# Token).}
\label{tab:focuscal_terms_vqa}
\end{table}
\begin{table}[t]
\centering
\label{tab:loss_comparison}

\begin{tabular}{lccc}
\toprule
\textbf{Dataset} & \textbf{BCE (\%)} & \textbf{MSE (\%)} & \makecell{\textbf{FocusCal (\%)}} \\
\midrule
MEDQA     & 84.2 & 83.1 & 88.8 \\
MMLU      & 85.0 & 86.7 & 92.4 \\
GSM8K     & 88.2 & 87.9 & 91.4 \\ \midrule
OKVQA     & 89.3 & 89.1 & 95.9 \\
VQA-v2       & 89.2 & 89.9 & 92.8 \\
ScienceQA & 91.4 & 91.8 & 93.4 \\
\bottomrule
\end{tabular}
\caption{Percentage of beneficial debate decisions using BCE or MSE compared to our FocusCal Loss.}
\label{tab:ablation_bce_vs_fmad}
\end{table}
\begin{table*}[!t]
\centering
\begin{tabular}{lcccccc}
\toprule
\multicolumn{1}{c}{\multirow{2.5}{*}{\textbf{Dataset}}}
& \multicolumn{2}{c}{\textbf{BCE}}
& \multicolumn{2}{c}{\textbf{MSE}}
& \multicolumn{2}{c}{\textbf{ECE}} \\
\cmidrule(lr){2-3} \cmidrule(lr){4-5} \cmidrule(lr){6-7}
& \textbf{Acc (\%)} & \textbf{\# Token}
& \textbf{Acc (\%)} & \textbf{\# Token}
& \textbf{Acc (\%)} & \textbf{\# Token} \\
\midrule
MEDQA     & 81.2 & 1,508 & \underline{81.4} & \underline{1,389} & \textbf{82.0} & \textbf{1,300} \\
MMLU      & \underline{89.0} & 1,296 & 88.8 & \textbf{1,005} & \textbf{89.2} & \underline{1,010} \\
GSM8K     & \underline{83.9} & 1,115 & 82.5 & \underline{1,099} & \textbf{84.8} & \textbf{1,025} \\ \midrule
OKVQA     & \underline{89.5} & 2,887 & 89.1 & \textbf{2,588} & \textbf{90.3} & \underline{2,601} \\
VQA-v2       & \underline{80.7} & \underline{3,626} & 79.8 & 3,798 & \textbf{81.3} & \textbf{3,489} \\
ScienceQA & \textbf{90.8} & 3,352 & 89.4 & \underline{2,899} & \textbf{90.8} & \textbf{2,893} \\
\midrule
Average   & \underline{85.9} & 2,297 & 85.2 & \underline{2,130} & \textbf{86.4} & \textbf{2,053} \\
\bottomrule
\end{tabular}
\caption{Accuracy (Acc) and average token cost (\# Token) per question when replacing the ECE term in FocusCal with BCE or MSE while maintaining $L_{\text{AF}}$ and $L_{\text{CP}}$. \textbf{Bold} values indicate the best result in each row, and \underline{underlined} values indicate the second best. }
\label{tab:ece_replacement_compact}
\end{table*}

\paragraph{FocusCal Loss Terms.}
In Table~\ref{tab:focuscal_terms_vqa}, we report accuracy (Acc) and total token cost (\#~Token) on the VQA-v2 dataset when using different subsets of the three components in FocusCal, i.e., $L_{\text{AF}}$, $L_{\text{CP}}$, and ECE. We report the results on the VQA-v2 dataset for brevity, since the other five datasets exhibit a similar performance pattern in both accuracy and token costs (e.g., full FocusCal consistently achieves higher accuracy than any two-term combination, which in turn outperforms any single-term loss). When we use only a single loss term, each component shows a distinct behavior. For example, using $L_{\text{CP}}$ alone yields the lowest token cost (3,379 tokens in total) but also the lowest accuracy of 78.1\%. Since $L_{\text{CP}}$ penalizes overconfident scores, it tends to push many predicted debate-triggering scores below the decision threshold. As a result, the classifier skips debate on more questions, including some that would have benefited from MAD, which reduces accuracy. Using ECE alone attains the best single-term accuracy of 79.1\% but with the highest token cost of 3,757 tokens. The ECE term adjusts the debate-triggering scores so that instances with scores close to the decision threshold (i.e., cases where the classifier is not clearly confident about skipping or triggering debate) are more likely to trigger MAD. This causes the classifier to trigger MAD on more questions, which improves accuracy but also increases token usage, even though some of these extra debates are unnecessary. Using $L_{\text{AF}}$ alone improves accuracy over $L_{\text{CP}}$ by emphasizing hard error cases (i.e., lowering the scores for wrong-but-confident answers). This behavior encourages the classifier to trigger debates precisely on those difficult cases, which helps correct more errors.

Combining two loss terms consistently improves performance over single-term configurations since each loss term targets a different failure mode of the debate decision classifier. On VQA-v2, the combination of $L_{\text{AF}}$ and $\text{ECE}$ and the combination of $L_{\text{CP}}$ and $\text{ECE}$ both raise accuracy to 79.7\% and 79.8\%, with token costs of 3,561 and 3,577, respectively. In these settings, ECE encourages the debate-triggering scores to match empirical correctness, while $L_{\text{AF}}$ focuses on overconfident errors and on examples near the decision threshold (where the classifier is unsure whether to skip or trigger debate), and $L_{\text{CP}}$ penalizes mismatches between the predicted score and the semantic uncertainty expressed in the self-critique response. 

For comparison, using $L_{\text{AF}}$ and $L_{\text{CP}}$ without ECE attains 79.7\% accuracy with 3,552 tokens, which is lower accuracy and slightly higher token cost than FocusCal. Removing the ECE term makes the debate-triggering scores less calibrated, so the classifier skips some recoverable errors (reducing accuracy) and also sends additional already correct or unrecoverable questions to debate (increasing token usage). Removing the ECE term makes the debate-triggering scores less calibrated, causing the classifier to skip some recoverable errors and thus reduce accuracy, and to trigger MAD on more already correct or unrecoverable instances and thus increase token costs.

Using all three terms together in FocusCal achieves the best overall performance on VQA-v2, reaching the highest accuracy of 81.3\% while reducing token usage to 3,489 tokens. In FocusCal, ECE calibrates the debate-triggering scores toward empirical correctness, $L_{\text{AF}}$ downweights wrong-but-confident predictions and highlights difficult cases, and $L_{\text{CP}}$ enforces consistency between predicted scores and semantic uncertainty cues in the self-critique. All three terms in FocusCal help the classifier identify more recoverable errors while avoiding unnecessary debates, which leads to higher accuracy and lower token usage.

\paragraph{FocusCal vs. BCE and MSE.} We compare FocusCal with two commonly used losses for training binary classifiers: BCE and MSE. Here, we focus on the quality of debate decisions rather than raw accuracy. Specifically, we measure the percentage of \emph{beneficial decisions} made by the classifier, defined as the percentage of examples where the classifier (\romannumeral1) correctly skips debate for questions that are already correct or not recoverable by MAD, or (\romannumeral2) correctly triggers debate for questions where MAD can improve the answer. 

As shown in Table~\ref{tab:ablation_bce_vs_fmad}, FocusCal achieves the highest rate of beneficial decisions across all datasets. For example, on OKVQA, FocusCal achieves 95.9\% beneficial decisions, while BCE and MSE achieve 89.3\% and 89.1\%, respectively. The improvement comes from the targeted design of FocusCal. By combining $L_{\text{AF}}$, $L_{\text{CP}}$, and ECE, FocusCal directly targets two key issues that matter for debate decisions: overconfident incorrect predictions and mismatches between the predicted debate-triggering scores and the uncertainty signals in the self-critique single-agent response. 
In contrast, BCE and MSE provide weaker guidance on when a debate is truly warranted, as they treat all classification errors uniformly and do not explicitly address these issues.

\begin{table*}[t]
\centering
\setlength{\tabcolsep}{1.8pt}
\begin{tabular}{lcccccccccccc}
\toprule
 & \multicolumn{4}{c}{\textbf{Single-Agent}} & \multicolumn{4}{c}{\textbf{Full-Debate MAD}}& \multicolumn{4}{c}{\textbf{Selective MAD}} \\
\cmidrule(lr){2-5} \cmidrule(lr){6-9}\cmidrule(lr){10-13}
\multirow{1}{*}{\textbf{Dataset}} & \multicolumn{2}{c}{\textbf{CoT}} & \multicolumn{2}{c}{\textbf{SC}} & \multicolumn{2}{c}{\textbf{MAD}}  & \multicolumn{2}{c}{\textbf{\textsc{GD}}} & \multicolumn{2}{c}{\textbf{\textsc{DOWN}}} & \multicolumn{2}{c}{\textbf{iMAD}} \\
\cmidrule(lr){2-3} \cmidrule(lr){4-5} \cmidrule(lr){6-7}\cmidrule(lr){8-9}\cmidrule(lr){10-11}\cmidrule(lr){12-13}
& \textbf{Acc (\%)} & \textbf{\# Token} & \textbf{Acc (\%)} & \textbf{\# Token} & \textbf{Acc (\%)} & \textbf{\# Token} & \textbf{Acc (\%)} & \textbf{\# Token} & \textbf{Acc (\%)} & \textbf{\# Token} & \textbf{Acc (\%)} & \textbf{\# Token} \\
\midrule
MEDQA     & 87.7 & 1,603 & 88.4 & 8,417  & \underline{91.3} & 46,352 & 90.4 & 160,232 & 89.0 & 2,084 & \textbf{91.6} & 2,165 \\
MMLU      & 85.1 & 1,482 & 86.5 & 8,351  & \underline{87.7} & 40,224 & \textbf{87.8} & 142,784 & 86.0 & 1,927 & \textbf{87.8} & 2,071 \\
GSM8K     & 70.4 & 1,550 & 73.6 & 8,925  & \underline{75.9} & 38,500 & 74.9 & 134,750 & 74.8 & 2,015 & \textbf{76.5} & 2,280 \\ \midrule
OKVQA     & 88.0 & 1,750 & 89.0 & 9,225  & 89.2 & 62,000 & \underline{89.3} & 215,000 & 88.5 & 2,275 & \textbf{89.8} & 2,400 \\
VQA-v2       & 76.7 & 1,661 & 77.2 & 9,536  & 78.3 & 41,119 & \underline{79.0} & 144,917 & 78.0 & 2,159 & \textbf{80.1} & 2,258 \\
ScienceQA & 86.8 & 1,600 & 86.8 & 8,002  & \underline{88.5} & 59,000 & 88.0 & 201,500 & 87.2 & 2,080 & \textbf{89.4} & 2,260 \\
\bottomrule
\end{tabular}
\caption{Performance on GPT-5 nano: Accuracy (Acc) and average token cost (\# Token) per question. \textbf{Bold} values indicate the best accuracy in each row, and \underline{underlined} values indicate the second best. }
\label{tab:fmad-performance-gpt5nano}
\end{table*}

\begin{table*}[t]
\centering
\setlength{\tabcolsep}{1.8pt}
\begin{tabular}{lcccccccccccc}
\toprule
 & \multicolumn{4}{c}{\textbf{Single-Agent}} & \multicolumn{4}{c}{\textbf{Full-Debate MAD}}& \multicolumn{4}{c}{\textbf{Selective MAD}} \\
\cmidrule(lr){2-5} \cmidrule(lr){6-9}\cmidrule(lr){10-13}
\multirow{1}{*}{\textbf{Dataset}} & \multicolumn{2}{c}{\textbf{CoT}} & \multicolumn{2}{c}{\textbf{SC}} & \multicolumn{2}{c}{\textbf{MAD}}  & \multicolumn{2}{c}{\textbf{\textsc{GD}}} & \multicolumn{2}{c}{\textbf{\textsc{DOWN}}} & \multicolumn{2}{c}{\textbf{iMAD}} \\
\cmidrule(lr){2-3} \cmidrule(lr){4-5} \cmidrule(lr){6-7}\cmidrule(lr){8-9}\cmidrule(lr){10-11}\cmidrule(lr){12-13}
& \textbf{Acc (\%)} & \textbf{\# Token} & \textbf{Acc (\%)} & \textbf{\# Token} & \textbf{Acc (\%)} & \textbf{\# Token} & \textbf{Acc (\%)} & \textbf{\# Token} & \textbf{Acc (\%)} & \textbf{\# Token} & \textbf{Acc (\%)} & \textbf{\# Token} \\
\midrule
MEDQA     & 80.3 & 1,030 & 81.0 & 5,465   & \underline{83.2} & 178,063 & 82.5 & 311,739 & 81.5 & 1,339 & \textbf{83.8} & 1,551 \\
MMLU      & 87.8 & 1,131 & 89.2 & 6,921   & \underline{91.3} & 149,686 & \underline{91.3} & 309,309 & 88.9 & 1,470 & \textbf{91.7} & 1,523 \\
GSM8K     & 74.2 & 1,099 & 75.0 & 5,290   & \underline{76.8} & 120,000 & \underline{76.8} & 284,000 & 75.5 & 1,274 & \textbf{77.6} & 1,366 \\ \midrule
OKVQA     & 88.5 & 1,000 & \underline{89.4} & 5,400   & 88.5 & 130,000 & 89.3 & 246,000 & 88.9 & 1,300 & \textbf{90.5} & 1,501 \\
VQA-v2       & 76.5 &   912 & 76.9 & 5,116   & 78.2 & 81,618  & \underline{79.0} & 159,560 & 76.8 & 1,186 & \textbf{79.4} & 1,350 \\
ScienceQA & 87.0 & 1.022 & 87.2 & 5,345   & \underline{90.5} & 115,000 & 89.0 & 283,000 & 87.5 & 1,287 & \textbf{90.9} & 1,338 \\
\bottomrule
\end{tabular}
\caption{Performance on Qwen 3.0: Accuracy (Acc) and average token cost (\# Token) per question. \textbf{Bold} values indicate the best accuracy in each row, and \underline{underlined} values indicate the second best. }
\label{tab:fmad-performance-qwen}
\end{table*}

\paragraph{Calibration Term of ECE Compared to MSE and BCE.}
To further evaluate the design choice of the calibration component ECE in FocusCal, we fix $L_{\text{AF}}$ and $L_{\text{CP}}$ and only replace the ECE term with BCE or MSE. As shown in Table~\ref{tab:ece_replacement_compact}, the model trained with ECE attains the best or tied accuracy on all datasets and usually uses fewer tokens. For example, on GSM8K, using ECE instead of MSE improves accuracy from 82.5\% to 84.8\% while reducing token usage by about 7\% (from 1,099 to 1,025 tokens per question). A similar pattern appears on VQA-v2, where ECE increases accuracy over BCE from 80.7\% to 81.3\% and lowers token usage by about 4\% (from 3,626 to 3,489 tokens). Averaged across all six datasets, training with ECE yields 86.4\% accuracy with 2,053 tokens per question, whereas BCE yields 85.9\% accuracy with 2,297 tokens, and MSE yields 85.2\% accuracy with 2,130 tokens. 

These results show that using ECE as the calibration component works better than using BCE or MSE in this setting. ECE groups instances into bins according to their predicted debate-triggering scores. For each bin, it compares the average predicted score with the actual fraction of instances in that bin where triggering debate is beneficial. When these two numbers differ a lot (e.g., when the bin has high scores but MAD rarely helps, or low scores but MAD often helps), the ECE term becomes large, pushing the model to adjust its scores in that bin. In contrast, BCE and MSE update each instance independently and do not explicitly fix these group-level mismatches. As a result, they are more likely to assign scores that are too high to many non-beneficial cases or too low to many beneficial cases, which makes the fixed decision threshold less reliable. 

%% file: Cross-model.tex
\subsection{Cross-LLM Evaluation}
\label{sec:cross-llm}
To evaluate the cross-model performance of \name, we conduct an additional cross-LLM study from three perspectives: (\romannumeral1) comparison of the token costs and accuracy of \name against all baselines when switching the base model to GPT-5 nano~\cite{gpt5} and Qwen 3.0~\cite{yang2025qwen3technicalreport}; and (\romannumeral2) analysis of how differences between base models influence the accuracy and token cost of \name.

\paragraph{Experimental Setup.}
We evaluate \name on the six benchmarks for both GPT-5 nano and Qwen 3.0, with detailed results reported in Table~\ref{tab:fmad-performance-gpt5nano} for GPT-5 nano and Table~\ref{tab:fmad-performance-qwen} for Qwen 3.0. All prompt templates, including the self-critique stage and memory synchronization, are kept identical to the main experimental setting with Gemini 2.0 Flash (Section~\ref{sec:evaluation}). For each base model, we first regenerate self-critique responses on the same two auxiliary datasets used in the main experiments (PubMedQA for QA and GQA for VQA), and retrain the debate decision classifier with FocusCal using these responses and their labels. We then select a decision threshold on these auxiliary datasets and keep both the classifier and the threshold fixed for that base model during evaluation. After this training step, we apply \name to all six evaluation datasets without any further tuning. For each method and each dataset, we report accuracy and the average total number of tokens per question (sum of input and output tokens) to evaluate performance.

\paragraph{Token Cost and Accuracy across Base Models.} 
Across both GPT-5 nano and Qwen 3.0, \name attains the highest accuracy on every dataset, while keeping token usage close to \textsc{DOWN} and far below full-debate methods. On GPT-5 nano with VQA-v2, \name achieves 80.1\% accuracy with 2,258 tokens per question, whereas \textsc{GD} reaches 79.0\% accuracy with 144,917 tokens, and MAD reaches 78.3\% accuracy with 41,119 tokens. Thus, \name matches or exceeds the accuracy of full-debate methods while using 98.4\% and 94.5\% fewer tokens than \textsc{GD} and MAD, respectively. On Qwen 3.0 with MMLU, \name attains 91.7\% accuracy with 1,523 tokens per question, while MAD and \textsc{GD} both reach 91.3\% accuracy but require 149,686 and 309,309 tokens. In this case, \name reduces token cost by 98.98\% relative to MAD and by 99.51\% relative to \textsc{GD}. Full-debate methods incur such high token costs because they trigger MAD on every instance, regardless of whether the instance is easy, already correctly answered, or unrecoverable. In every round, they append new messages to the shared transcript and resend the entire cumulative transcript to all agents. This repeated rebroadcast quickly increases the number of input token costs, leading to very large token budgets even on instances where debate does not change the final answer. In contrast, \name triggers debate only when the debate decision classifier detects hesitation or conflict in the self-critique single-agent response. As a result, extra tokens are spent mainly on the questions where triggering MAD is likely to correct an error.

Compared with \textsc{DOWN}, \name consistently achieves higher accuracy with only a slight increase in token usage on both base models. On GPT-5 nano, for example, \name improves MEDQA accuracy from 89.0\% to 91.6\% while using 81 more tokens per question (about 3.9\% more tokens). On Qwen 3.0, \name improves MMLU accuracy from 88.9\% to 91.7\% with 53 additional tokens per question (about 3.6\% more tokens). These results are consistent with the main experiments on the Gemini 2.0 Flash model and show that the advantage of \name over \textsc{DOWN} persists across base models. These gains arise from the different ways the two methods decide when to trigger debate. \textsc{DOWN} relies solely on the model-generated confidence score in the single-agent answer and compares this score against a dataset-specific threshold. As confidence scores from LLMs are often overconfident even when the answer is incorrect, \textsc{DOWN} frequently skips debate on instances where additional debate would be beneficial. In contrast, \name uses a dedicated debate decision classifier trained with the FocusCal loss to decide whether to trigger MAD. This classifier learns to trigger MAD when the self-critique shows hesitation or conflicting reasoning, and to skip debate when the self-critique strongly supports the initial correct answer or indicates that the error is unlikely to be fixed by debate. As a result, \name spends additional tokens mainly on instances where triggering MAD is likely to correct the error, which explains why it achieves higher accuracy than \textsc{DOWN} with only a slightly higher token usage.

Furthermore, on GSM8K, a reasoning-heavy math word-problem dataset that requires multi-step numerical reasoning, \name improves over single-agent baselines on both GPT-5 nano and Qwen 3.0 while adding only a small number of extra tokens per question. In these settings, many errors arise from intermediate reasoning steps. The self-critique stage therefore elicits additional reasoning that reveals uncertainty and alternative reasoning paths, which gives the debate decision classifier clearer signals about when triggering MAD is likely to correct the error. For VQA or short factual tasks such as OKVQA, VQA-v2, and MMLU, the accuracy gains are smaller because these tasks often hinge on a single decisive visual cue or fact. When that key cue is hard to obtain, triggering MAD adds little new information and has limited ability to correct the error.

\paragraph{Effect of Base Model Differences.}
Under the same prompts and overall experimental setup, we observe that \name uses fewer tokens per question on Qwen 3.0 than on GPT-5 nano (see Table~\ref{tab:fmad-performance-qwen} for Qwen 3.0 and Table~\ref{tab:fmad-performance-gpt5nano} for GPT-5 nano). This difference reflects the combined effect of each model’s tokenizer and how long the models typically respond under the same prompts. As a result, \name achieves similar accuracy on both models while Qwen 3.0 uses fewer tokens per question than GPT-5 nano. This suggests that \name benefits more from base models whose responses are compact but still informative. In this case, fewer tokens are spent per question, while the debate decision classifier still receives enough useful signals, such as uncertainty or conflicting reasoning, to decide whether triggering MAD is beneficial.

Overall, these results show that, for each base model, the debate decision classifier trained only once on the auxiliary datasets (e.g., PubMedQA and GQA) generalizes well to the six evaluation datasets. Across both GPT-5 nano and Qwen 3.0, \name consistently improves accuracy over single-agent baselines and over \textsc{DOWN}, while avoiding the heavy token cost of full-debate methods.

%% file: 7_Discussion.tex
\section{Discussion}
\label{sec:discussion}

While \name demonstrates its strong improvements in both token efficiency and accuracy, we discuss its design assumptions (i.e., limited labeled data, prompt engineering, and label design choices for the debate classifier) and potential extensions for future work.

\paragraph{Limited Labeled Data.}
One key requirement for \name is constructing a training dataset with correctness labels for single-agent answers. 
We construct this dataset by running the self-critique single-agent pipeline on two auxiliary datasets, PubMedQA (a QA dataset) and GQA (a VQA dataset), and manually checking whether the chosen option matches the ground-truth answer. We label the single-agent answer as correct when its chosen option matches the ground-truth answer, and as incorrect otherwise. These binary correctness labels serve as supervision for training the debate decision classifier: each example is paired with features extracted from the self-critique response, and the classifier learns to predict whether a debate would be beneficial. This labeling process is lightweight and uses only two datasets, and our careful dataset selection exposes the classifier to diverse reasoning behaviors. This design allows the classifier to generalize well and perform effectively on six held-out datasets without further tuning. However, since the classifier is trained offline and remains fixed during deployment, it may not effectively adapt to changes in model behaviors. Future work could explore online adaptation through feedback-driven updates or weak supervision to further improve generalization.

Although the labeling relies on only two datasets, they cover diverse question types and reasoning patterns, which helps the classifier generalize to six held-out evaluation datasets without additional tuning. A limitation is that the classifier is trained offline and kept fixed at deployment, so it cannot adapt when the underlying model behaviors drift over time or when the deployment domain changes substantially. Future work could explore online adaptation, for example, through feedback-driven updates or weak supervision, to further improve generalization.

\paragraph{Prompt Engineering.}
Another consideration lies in the reliance on structured self-critique prompting, which \name uses to extract interpretable features from the model’s output to train the downstream debate decision classifier to capture general model behaviors. However, this design assumes that the underlying LLM agent can effectively articulate its reasoning and express uncertainty. In domains where the model’s responses are incoherent or fail to convey uncertainty clearly, the extracted features may become less informative. In such cases, additional prompt refinement may be necessary to preserve debate decision quality.

\paragraph{Label Design Choices for the Debate Classifier.} 
For the debate decision classifier, we explored two label design choices. The first, which we adopt in \name, is a simple correctness-based labeling strategy. We construct a training dataset by running the self-critique single-agent pipeline on two auxiliary datasets (PubMedQA for QA and GQA for VQA) and comparing the single-agent answer with the ground-truth label for each instance, as described in Section~\ref{sec:overview}. The second strategy is a more direct debate-benefit labeling scheme that labels each instance according to whether MAD actually improves the single-agent answer. In this debate-benefit scheme, we run both the single-agent and MAD pipelines on the training instances and assign a binary label $y$ to each instance. We set $y = 1$ if the single-agent answer is wrong and the MAD answer is correct, so debate should be triggered. In all other cases, including when both answers are correct, both are wrong, or the single-agent answer is correct, but the MAD answer is wrong, we set $y = 0$, since debate is unnecessary or harmful.

In practice, this debate-benefit labeling scheme performed poorly for training the debate decision classifier, for two main reasons. First, the positive cases with $y = 1$, where MAD actually fixes a wrong single-agent answer, are much fewer than the negative cases (about 6-9\% of all instances), so the training dataset is highly imbalanced. The classifier therefore tends to learn a trivial policy that rarely triggers debate. Second, the labels produced by this scheme do not align well with what the classifier actually observes. In the debate-benefit scheme, each label records whether MAD corrects the single-agent error in that particular run. However, MAD's success depends not only on the single-agent output, but also on the detailed debate trajectory and sampling randomness during generation. For the same question and an almost identical self-critique single-agent response, one MAD run may fix the answer while another may fail. The debate decision classifier, by design, only uses features derived from the single-agent self-critique and never has access to the MAD trajectory when it makes its prediction. Consequently, two training instances with nearly identical single-agent features can receive opposite labels, depending on how MAD happens to behave in that run. From the classifier's perspective, the same input pattern is sometimes labeled ``trigger'' and sometimes labeled ``skip''. As a result, the labels are noisy and inconsistent with the input features, and the classifier cannot learn a clear rule for when triggering MAD is beneficial.

Overall, these observations lead us to adopt the correctness-based labeling strategy in our final design: it is both cheaper to obtain and more robust for training the debate decision classifier. In contrast, constructing debate-benefit labels requires running the full MAD pipeline for every training example, which is far more expensive in token cost than the correctness-based labeling pipeline. In \name, the training labels depend only on whether the single-agent answer is correct or not. Hence, they are robust and do not involve any randomness from the debate process during labeling. Randomness in MAD appears only at inference time and does not affect how the classifier is trained. Training with binary correctness labels, together with the FocusCal loss, encourages the classifier to trigger MAD mainly on instances where the self-critique single-agent response reveals high uncertainty or self-contradictory. These cases strongly indicate single-agent errors that MAD can often correct. This does not mean that every incorrect single-agent answer will trigger MAD. Instead, the classifier focuses on those errors that can be detected from the self-critique response and for which debate is more likely to be beneficial.

\paragraph{Future Work.}
An interesting extension is to make the debate-triggering decision during output generation rather than waiting for the entire self-critique response to be generated. In our current deployment, the LLM agent is accessed only through a black-box API (e.g., Gemini or GPT), so \name can decide whether to trigger debate only after the complete self-critique response is generated. With open-source models or streaming APIs, it would be possible to monitor token-by-token generation and detect early signs of either hesitation or strong confidence in the emerging rationale. Such signals could be used to trigger MAD early when needed or skip debate when the answer appears well justified, potentially reducing both inference time and token costs while maintaining accuracy. A further direction is to combine such in-generation signals with internal model states (e.g., logit trajectories or entropy over the candidate answers) and learn an adaptive policy (e.g., via reinforcement learning) that allocates debate budget dynamically over the course of generation.